\documentclass[acmsmall,screen,nonacm]{acmart}
\usepackage{multirow}
\usepackage[most]{tcolorbox}
\usepackage{tikz}
\usepackage[edges]{forest}
\definecolor{hidden-draw}{RGB}{205, 44, 36}
\definecolor{hidden-blue}{RGB}{194,232,247}
\definecolor{hidden-orange}{RGB}{243,202,120}
\definecolor{hidden-yellow}{RGB}{242,244,193}
\definecolor{tree-level-1}{RGB}{245,20,85}
\definecolor{tree-level-2}{RGB}{246,86,118}
\definecolor{tree-level-3}{RGB}{248,177,193}
\definecolor{tree-leaf}{RGB}{176,230,198}
\definecolor{crimson}{HTML}{DC143C}
\definecolor{ggreen}{HTML}{1b9e77}
\definecolor{Self}{RGB}{255,0,128}
\definecolor{Ensemble}{RGB}{0,127,255}
\definecolor{Iterative}{RGB}{153,51,255}

\definecolor{exemplar1}{RGB}{136,98,148}
\definecolor{exemplar2}{RGB}{148,210,242}
\definecolor{knowledge1}{RGB}{249,219,152}
\definecolor{knowledge2}{RGB}{255,245,220}

\usepackage{subcaption}
\usepackage{pgfplots}
\pgfplotsset{compat=1.17}
\usetikzlibrary{pgfplots.groupplots}
\definecolor{darkpastelgreen}{rgb}{0.01, 0.75, 0.24}
\definecolor{wingreen}{rgb}{0,0.45,0.24}
\definecolor{losered}{rgb}{1.0,0.1,0.24}
\definecolor{awesome}{rgb}{1.0, 0.13, 0.32}
\definecolor{lightcoral}{rgb}{0.94, 0.5, 0.5}
\definecolor{lightgreen}{rgb}{0.56, 0.93, 0.56}
\definecolor{harvestgold}{rgb}{0.85, 0.57, 0.0}
\definecolor{brightlavender}{rgb}{0.75, 0.58, 0.89}
\definecolor{capri}{rgb}{0.0, 0.75, 1.0}
\definecolor{carminepink}{rgb}{0.92, 0.3, 0.26}
\definecolor{celadon}{rgb}{0.67, 0.88, 0.69}
\definecolor{darkpastelgreen}{rgb}{0.01, 0.75, 0.24}

\AtBeginDocument{%
  \providecommand\BibTeX{{%
    \normalfont B\kern-0.5em{\scshape i\kern-0.25em b}\kern-0.8em\TeX}}}

\setcopyright{acmlicensed}
\copyrightyear{2024}
\acmYear{2024}
\acmDOI{XXXXXXX.XXXXXXX}

\begin{document}

\title{
A Systematic Survey of Text Summarization: From Statistical Methods to Large Language Models
}

\author{Haopeng Zhang}
\email{haopengz@hawaii.edu}
\affiliation{%
  \institution{University of Hawaii, Manoa}
  \city{Honolulu}
  \country{USA}
}

\author{Philip S. Yu}
\email{psyu@uic.edu}
\affiliation{%
  \institution{University of Illinois at Chicago}
  \city{Chicago}
  \country{USA}
}

\author{Jiawei Zhang}
\email{jiawei@ifmlab.org}
\affiliation{%
  \institution{University of California, Davis}
  \city{Davis}
  \country{USA}
}
\renewcommand{\shortauthors}{Haopeng, et al.}

\begin{abstract}

Text summarization research has undergone several significant transformations with the advent of deep neural networks, pre-trained language models (PLMs), and recent large language models (LLMs). This survey thus provides a comprehensive review of the research progress and evolution in text summarization through the lens of these paradigm shifts. It is organized into two main parts: (1) a detailed overview of datasets, evaluation metrics, and summarization methods before the LLM era, encompassing traditional statistical methods, deep learning approaches, and PLM fine-tuning techniques, and (2) the first detailed examination of recent advancements in benchmarking, modeling, and evaluating summarization in the LLM era. By synthesizing existing literature and presenting a cohesive overview, this survey also discusses research trends, open challenges, and proposes promising research directions in summarization, aiming to guide researchers through the evolving landscape of summarization research.
\end{abstract}




\maketitle

\section{Introduction}

Text summarization is one of the most critical and challenging tasks in Natural Language Processing (NLP). It is defined as the process of distilling the most important information from a source (or sources) to produce an abridged version for a particular user (or users) and task (or tasks)~\cite{mani1999advances}. With the explosion in the amount of textual information available online since the advent of the Internet, summarization research has garnered significant attention.

\begin{figure}[!t]
    \centering
    \includegraphics[width=\textwidth]{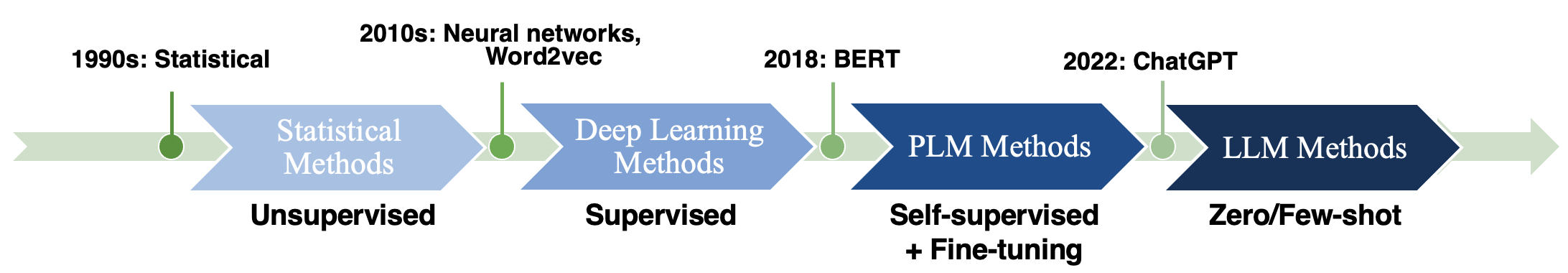}
    \caption{The Evolution of the Four Major Paradigms in Text Summarization Research.
    }
    \label{fig:timeline}
\end{figure}

Early efforts in building automatic text summarization (ATS) systems date back to the 1950s~\cite{luhn1958automatic}. Subsequently, unsupervised feature-based systems emerged with advances in statistical machine learning during the 1990s and early 2000s~\cite{carbonell1998use,mihalcea2004textrank,erkan2004lexrank}. In the 2010s, the focus of summarization research shifted towards training deep learning frameworks in a supervised manner, leveraging the availability of large-scale training data~\cite{nallapati2017summarunner,cheng2016neural,see2017get,narayan2018ranking}. More recently, the advent of self-supervised pre-trained language models (PLMs) such as BERT~\cite{devlin2018bert} and T5~\cite{raffel2020exploring} has significantly enhanced summarization performance through the `pre-train, then fine-tune' pipeline~\cite{liu2019text,zhong2020extractive,liu2022brio}. This progression culminates in the current era dominated by large language models (LLMs)~\cite{brown2020language,achiam2023gpt}. Looking back at the development history of summarization approaches, we can generally categorize it into four stages based on the underlying paradigms: the statistical stage, the deep learning stage, the pre-trained language model fine-tuning stage, and the current large language model stage, as illustrated in Fig.~\ref{fig:timeline}.

Recently, the emergence of LLMs has revolutionized both academic NLP research and industrial products due to their remarkable ability to understand, analyze, and generate texts with vast amounts of pre-trained knowledge. By leveraging extensive corpora of text data, LLMs can capture complex linguistic patterns, semantic relationships, and contextual cues, enabling them to produce high-quality summaries that rival those crafted by humans~\cite{zhang2024benchmarking}. There is no doubt that LLMs have propelled the field of summarization into a new era~\cite{achiam2023gpt,team2023gemini,touvron2023llama}.

At the same time, many NLP researchers are experiencing an existential crisis triggered by the astonishing success of LLM systems. Text summarization is definitely one of the most affected fields in which researchers argue that summarization is (almost) dead~\cite{pu2023summarization}. After such a disruptive change to our understanding of the summarization field, what is left to do? The ever-evolving landscape of LLMs, characterized by the continual development of larger, more powerful models, poses both opportunities and challenges for researchers and practitioners in the field of summarization.

This survey paper aims to provide a comprehensive overview of the state-of-the-art summarization research endeavors in the new era of LLMs. We first categorize existing approaches and discuss the problem formulation, evaluation metrics, and commonly used datasets. Next, we systematically analyze representative text summarization methods prior to the LLM era,  encompassing traditional statistical methods, deep learning approaches, and PLM fine-tuning techniques. Furthermore, we synthesize insights from recent summarization literature during the LLM era, discuss research trends and open challenges, and propose promising research directions in summarization. This survey seeks to foster a deeper understanding of the advancements, challenges, and future prospects in leveraging LLMs for text summarization, ultimately contributing to the ongoing development and refinement of NLP research.

\subsection{Major Differences}

Several surveys have been conducted to examine various aspects of summarization. Nevertheless, existing surveys mostly focus on investigating traditional statistical approaches and deep learning-based summarization approaches~\cite{dong2018survey,el2021automatic,cajueiro2023comprehensive}. The scarcity of comprehensive up-to-date survey studies and the lack of consensus continue to inhibit progress. With the significant paradigm shift initiated by pre-trained language models and, more recently, large language models, there is still a dearth of thorough investigations that comprehensively encompass the continuous progress in the field of summarization in this new era.

For example, early survey papers~\cite{el2021automatic,mridha2021survey} provide a comprehensive survey of statistical and deep learning-based automatic text summarization models, discussing their detailed taxonomy and applications. Following this, ~\citet{cajueiro2023comprehensive} provide a comprehensive literature review of the methods, data, evaluation, and coding of ATS up to the advent of pre-trained language models.

Researchers have also produced surveys with a more specific focus.  Survey paper~\cite{koh2022empirical} focuses on the datasets, methods, and evaluation metrics of long documents such as academic articles and business reports. Another work~\cite{salchner2022survey} focuses on graph neural network-based (GNN-based) approaches to the task of automatic text summarization. \citet{cao2022survey} conducts a survey on neural abstractive summarization methods and explores the factual consistency of the abstractive summarization systems.

Considering the rapid advancement of summarization approaches and the disruptive change introduced by LLMs, we believe it is imperative to review the details of representative methods both before and during the LLM era, analyze the uniqueness of each method, and discuss open challenges and prospective directions to facilitate the further advancement of the field.

\subsection{Main Contributions}

The main contributions of this survey are to investigate text summarization approaches through the lens of paradigm shifts and to review recent research endeavors in the era of LLMs. Detailed contributions of this survey paper include:

\begin{itemize}
\item This survey paper serves as the first comprehensive study of the task of text summarization in the new era of LLMs. We propose the first taxonomy of LLM-based summarization literature in Figure~\ref{fig:tax}, based on objectives and methodology. The necessary detailed information on the research efforts, including benchmarking studies, modeling studies, and LLM-based summary evaluation studies, is also presented in Tables~\ref{tab:llm_bench}, \ref{tab:llm_method}, and \ref{tab:llm_evaluation}.

\item In addition, this survey proposes a new categorization of summarization methods based on the underlying paradigms. We also present a comprehensive investigation of representative summarization algorithms and approaches before the LLM era – from traditional statistical models to deep learning models to PLM fine-tuning methods. The taxonomy of these works is presented in Figure~\ref{fig:pre_tax}, and detailed information is shown in Table~\ref{tab:method}.

\item Moreover, we discuss the categorization of summarization methods, summarize commonly used datasets for text summarization with URL links in Table~\ref{tab:dataset}, review popular summary evaluation metrics, and present quantitative result comparison on the CNN/DM dataset in Table~\ref{tab:result}.

\item Finally, we analyze the underlying trends in summarization research, discuss unresolved challenges within the field, and delineate prospective directions for investigation in the new era of LLMs to foster further advancement.

\end{itemize}

\subsection{Organization}

This survey is organized as follows: Section~\ref{sec:bg} provides the background of text summarization, including categorizations of approaches, problem formulation, evaluation metrics, and common datasets. Section~\ref{sec:prior} reviews prominent summarization methods before the era of LLMs, encompassing statistical approaches, deep learning-based approaches, and PLM fine-tuning approaches. Section~\ref{sec:llm} explores recent advancements in benchmarking LLMs for summarization (\S\ref{sec:benchmark}), developing LLM-based summarization systems (\S\ref{sec:method}), and evaluating summaries with LLMs (\S\ref{sec:evaluation}). Finally, Section~\ref{sec:future} discusses open problems and future research directions in summarization.

\section{Background}
\label{sec:bg}

This section provides essential background information on text summarization. Firstly, we outline the general categorization of text summarization tasks in Fig.~\ref{fig:category} and discuss the typical problem formulations of extractive and abstractive summarization. Additionally, we review common evaluation metrics used to assess summarization performance before the emergence of LLMs. Finally, we offer an overview of widely utilized benchmark datasets for summarization, as detailed in Table~\ref{tab:dataset}.


\subsection{Categorization}

Text summarization is the task of creating a short, accurate, and fluent summary of a longer text document. It finds applications across various domains such as news aggregation, document summarization, social media analysis, and more. Its primary goal is to help users quickly grasp the main points of a document or a piece of content without having to read through the entire text. However, text summarization faces challenges such as maintaining coherence and preserving important details, dealing with various types of content, and ensuring the summary is both accurate and concise. 

\begin{figure}[!tbp]
    \centering
    \includegraphics[width=0.9\textwidth]{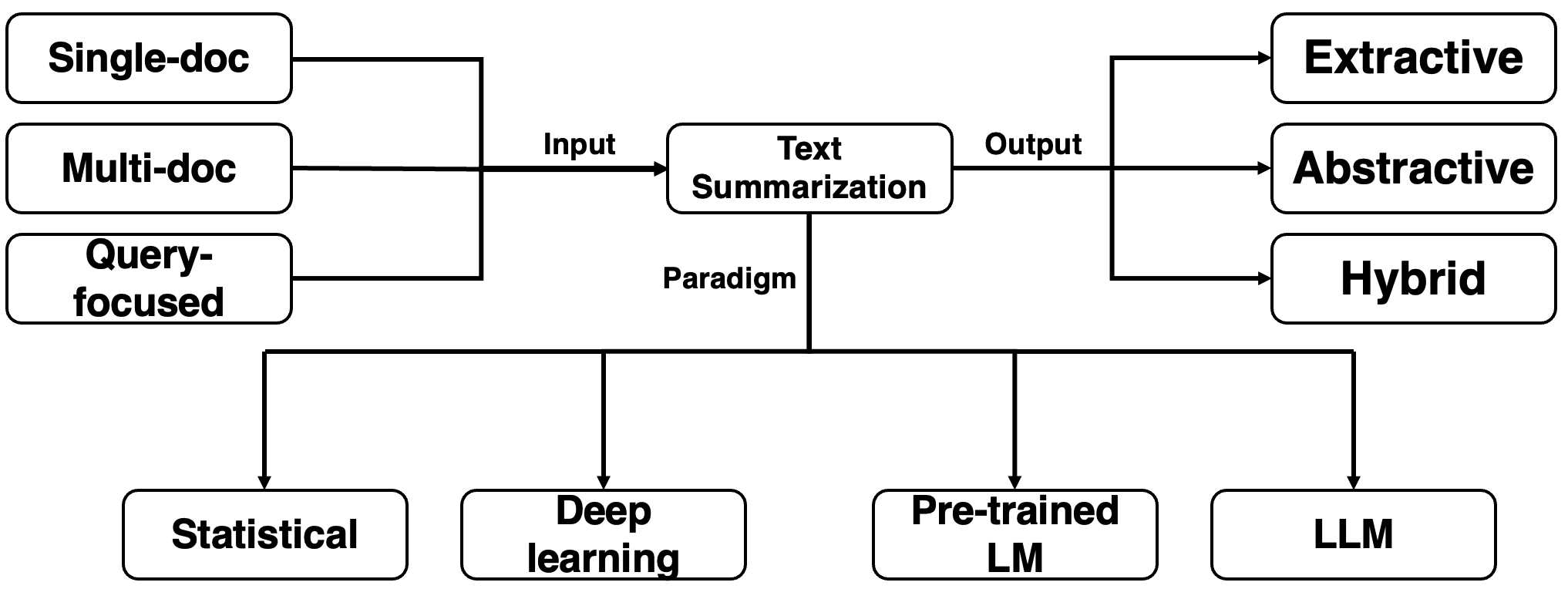}
    \caption{Categorization of Summarization Approaches based on input formats and output styles.}
       \label{fig:category}
\end{figure}

As depicted in Fig.~\ref{fig:category}, text summarization approaches can be classified into various groups depending on the format of the input document, the style of the summary output, and the underlying paradigm.

\subsubsection{{Input: Single-document vs. Multi-document vs. Query-focused}}

 Common text summarization approaches can be categorized based on the format of the input source document into single-document summarization (SDS), multi-document summarization (MDS), and query-focused summarization (QFS). SDS summarizes a single article~\cite{liu2019text}, while MDS takes a cluster of documents, typically on the same topic, as input~\cite{fabbri2019multi}. QFS, on the other hand, aims to generate a summary that specifically addresses an input query, such as a topic, keywords, or an entity~\cite{zhong2021qmsum,bahrainian2022newts}. Additionally, some summarization tasks involve multilingual inputs (e.g., translating from Chinese to English)~\cite{hasan2021xl} and multimodal inputs (e.g., combining text and images)~\cite{jiang2023exploiting,lin2023videoxum}.

\subsubsection{Output: Extractive vs. Abstractive vs. Hybrid} 

Text summarization approaches can also be classified into extractive, abstractive, and hybrid methods based on how the output summary is generated. As shown in the examples in Fig.~\ref{fig:example}, extractive methods create summaries by extracting sentences from the original documents~\cite{liu2019text}. Abstractive methods generate the summary word by word with novel content~\cite{lewis2019bart}. Hybrid methods combine both extraction and abstraction techniques~\cite{dou2020gsum}.

Specifically, \textit{extractive summarization} systems aim to identify and select significant text spans and sentences from the original document to form the summary. These summaries are inherently faithful, fluent, and accurate, but they may suffer from issues such as redundancy and incoherence. Statistical and deep learning-based summarization systems~\cite{mihalcea2004textrank,erkan2004lexrank,nallapati2017summarunner} are mostly extractive and typically frame extractive summarization as a sequence labeling and ranking problem to identify the most salient sentences.

\textit{Abstractive summarization} systems, in contrast, generate summaries from scratch, similar to how humans write summaries. These systems produce more flexible and concise summaries but may encounter problems like hallucination and unfaithfulness. Abstractive summarization is often formulated as a sequence-to-sequence (seq2seq) problem~\cite{sutskever2014sequence} and utilizes encoder-decoder frameworks~\cite{lewis2019bart,zhang2020pegasus} or autoregressive language models as in~\cite{brown2020language}.

\textit{Hybrid summarization} approaches attempt to combine the strengths of both extractive and abstractive methods to achieve more balanced results~\cite{dou2020gsum,zhang2023extractive}. These systems typically extract key information from the source document first and then use this extracted information to guide the abstractive summary generation process.

\begin{figure}[t]
    \begin{tcolorbox}[fontupper=\scriptsize\sffamily, fontlower=\scriptsize\sffamily, colback=gray!5, left=1mm, right=1mm, top=1mm, bottom=1mm]
        \textbf{Document: }Daredevil Nik Wallenda says he’ll walk untethered on top of a 400-foot observation wheel in
Orlando, Florida, this month. Wallenda said Monday at a New York City news conference that
the Orlando Eye will be moving when he attempts his feat on April 29. The Orlando Eye, part
of a new entertainment complex, will offer views of central Florida from inside 30 enclosed,
air-conditioned glass capsules when it opens to the public on May 4. Eyes on the prize: high-wire
performer Nik Wallenda announces his latest stunt at the 400-foot Orlando Eye, during a news
conference, in New York on Monday. {\color{crimson}Tough challenge: the 36-year-old daredevil will walk atop
the Orlando Eye as it turns on April 29.} The Orlando Eye team issued a statement saying it’s
excited to have Wallenda attempt the ‘amazing stunt.’ No distance for the performance has been
set yet, but Wallenda, 36, said he was not likely to walk the entire 20 minutes or so that it takes
the wheel to go a full circle. Wallenda previously walked atop a Ferris wheel in Santa Cruz,
California, but he said the size of the much larger Orlando wheel and the fact that he will not use
a pole sets this stunt apart. {\color{crimson}The seventh-generation member of the ‘Flying Wallenda’ family of
acrobats has walked across the Grand Canyon and Niagara Falls.} \color{crimson}{In November, he walked twice
between two Chicago skyscrapers without a net or harness, doing the second walk blindfolded.}
Wallenda is the great-grandson of Karl Wallenda, who fell to his death in Puerto Rico at 73.
        \tcblower
        \textbf{Reference Summary: }The 36-year-old will stage his next stunt on April 29. In November, Wallenda walked back and
forth between two Chicago skyscrapers in a live television event. His great-grandfather Karl
Wallenda died in a tightrope walk in Puerto Rico in 1978. Wallenda has also tightrope walked
across Niagara Falls and the Grand Canyon. \\
        \textbf{Extractive Summary: }Tough challenge: the 36-year-old daredevil will walk atop the Orlando Eye as it turns on April
29. The seventh-generation member of the ‘Flying Wallenda’ family of acrobats has walked
across the Grand Canyon and Niagara Falls. In November, he walked twice between two Chicago
skyscrapers without a net or harness, doing the second walk blindfolded.        \\
\hfill
        \textbf{Abstractive Summary: }aredevil Nik Wallenda plans to walk untethered atop the 400-foot Orlando Eye observation
wheel on April 29th. The 36-year-old high-wire performer, a seventh-generation member of
the "Flying Wallenda" acrobat family, has previously completed daring walks across the Grand
Canyon, Niagara Falls, and between Chicago skyscrapers. Although the exact distance for this
stunt hasn’t been determined, Wallenda stated he’s unlikely to walk the entire 20 minutes it takes
for the wheel to complete a full circle.
    \end{tcolorbox}
        \caption{Example of Abstractive and Extractive Summaries for a News Article from CNN/Dailymail.}
    \label{fig:example}
\end{figure}

\subsubsection{Backbone Paradigm: Statistical vs. Deep learning vs. PLM fine-tuning vs. LLM}

 As previously mentioned, the development of text summarization can be categorized into four major stages based on the underlying paradigms: the statistical stage, the deep learning stage, the pre-trained language model fine-tuning stage, and the current large language model stage.

\textit{Statistical Stage} features unsupervised methods for summarization, including heuristic-based approaches~\cite{carbonell1998use}, optimization-based approaches~\cite{mcdonald2007study,lin2011class}, and graph-based approaches~\cite{mihalcea2004textrank,erkan2004lexrank}. Hand-crafted features and frequency-based features like the term frequency-inverse document frequency (TF-IDF) are commonly used~\cite{edmundson1969new}.

\textit{Deep Learning Stage} approaches rely on domain-specific training data in the form of source document-summary pairs to train neural networks in a supervised manner~\cite{nallapati2017summarunner,cheng2016neural,see2017get,narayan2018ranking}. Word embedding techniques~\cite{mikolov2013efficient,pennington2014glove} are also commonly used. During this stage, researchers introduced various training corpora to advance summarization development~\cite{hermann2015teaching,narayan2018don,sandhaus2008new}.

\textit{PLM Fine-tuning Stage}: This stage relies on advances in large-scale, self-supervised pre-trained language models~\cite{lewis2019bart,liu2019roberta}. The introduction of Bidirectional Encoder Representations from Transformers (BERT)~\cite{devlin2018bert} marks the beginning of this stage, resulting in significant performance improvements. Pre-training with large amounts of text data equips PLMs with language patterns and parametric knowledge, which are important for document comprehension and text generation, thus benefiting downstream tasks. This stage features a `pre-train, then fine-tune' pipeline that fine-tunes pre-trained language models on task-specific data to further improve performance on downstream tasks~\cite{liu2019text}.

\textit{LLM Stage}: Most recently, advances in large language models have reshaped summarization research. The strong understanding and instruction-following capabilities of these models have enabled the development of zero-shot and few-shot summarization systems, bringing new opportunities in this emerging era~\cite{brown2020language,wei2022chain}. The beginning of this stage is marked by the introduction of Generative Pre-trained Transformer 3 (GPT-3)~\cite{brown2020language} by OpenAI, with a model size of $175$ billion parameters and strong few-shot capabilities.

\subsection{Problem Formulation}
\subsubsection{Extractive Summarization Formulation}

Extractive summarization produces a summary by identifying and directly extracting key sentences from a source document. Without loss of generality, we present the problem formulation of extractive summarization for a single document $D$ here.

Formally, given a document with $n$ sentences as $D=\{s_1^d, s_2^d, ..., s_n^d\}$, extractive summarization systems aim to form a summary $S=\{s_1^s, s_2^s, ..., s_m^s \}$ with $m\:(m \ll n)$ sentences by directly extracting sentences from the source document. Most existing approaches formulate extractive summarization as a binary sequence labeling problem and assign each sentence a $\{0,1\}$ label. Here, label $1$ indicates that the sentence is salient and will be included in summary $S$, while label $0$ indicates that the sentence is not salient and will be ignored. 

However, extractive ground truth in the form of sentence-level binary labels (ORACLE) is rarely available since most existing benchmark datasets use human-written summaries as gold standards. Therefore, it is common to use a greedy algorithm to generate a sub-optimal ORACLE, consisting of multiple sentences that maximize the ROUGE-2 score~\cite{lin2003automatic} against the reference summary, as described by \cite{nallapati2017summarunner}. Specifically, sentences are incrementally added to the summary one at a time so that the ROUGE score of the current set of selected sentences is maximized with respect to the entire gold summary. This process continues until no remaining candidate sentence improves the ROUGE score upon addition to the current summary set.


\subsubsection{Abstractive Summarization Formulation}
Abstractive summarization is commonly framed as a sequence-to-sequence problem, typically tackled using the encoder-decoder neural architecture. In this setup, the encoder processes the source document $D$ into continuous vector representations. Subsequently, the decoder uses these representations to generate the summary word by word in an autoregressive manner. Most abstractive summarization systems adopt sequential models and utilize a teacher-forcing training strategy~\cite{vaswani2017attention}.

Formally, abstractive summarization systems aim to generate a summary $S$ word by word, conditioned on the corresponding source document $D$. Model parameters $\theta$ are trained to maximize the conditional likelihood of the outputs:
$$
    \arg\max_{\theta} \log p ({s}^i \,|\, D; \theta),
$$
where ${s}^i$ represents the i-th word in the generated summary $S$.

\subsubsection{Hybrid}

Hybrid summarization approaches typically adopt an extract-then-generate pipeline, where key information $G$ is initially extracted from the source document before being used to guide the decoding process. Formally, it can be viewed as augmenting the abstractive model by incorporating an additional signal $G$, alongside the source document ${S}$, to guide the generation process:
$$
\arg\max_{\theta} \log p ({s}^i \,|\, D, G; \theta).
$$

Commonly employed guiding signals $G$ include extracted summaries, keywords, relations, and information retrieved from external sources~\cite{zhong2020extractive,dou2020gsum,pilault2020extractive,wang2022salience}.

\subsection{Evaluation Metrics}
\label{metric}

Early research focuses on evaluating the quality of summaries using human judges~\cite{dang2008overview, mani1999advances, nenkova2004evaluating}. Since manually evaluating generated summaries is both costly and impractical, researchers have developed automatic metrics instead. Generally, the overall quality of a summary is evaluated along four dimensions: coherence, consistency, fluency, and relevance~\cite{zhong2022towards}. Recently, \citet{fabbri2021summeval} conduct a comprehensive meta-evaluation of common automatic evaluation metrics, incorporating crowd-sourced human annotations.

\subsubsection{Similarity-based Summary Evaluation}

Existing summary evaluation metrics primarily rely on the similarity between the generated summary and a reference gold summary (human-written) as the main criterion.

ROUGE F-scores~\cite{lin2003automatic} have long been the standard way to evaluate summarization model performance. They measure the n-gram lexical overlap between the reference and candidate summaries. Specifically, ROUGE-1 and ROUGE-2 scores refer to unigram and bigram overlap, respectively, indicating summary informativeness. The ROUGE-L score, which refers to the longest common sequence, indicates summary fluency. However, since ROUGE is based on exact n-gram matches, it ignores overlaps between synonymous phrases and penalizes models that generate novel wordings and phrases.

Researchers have also explored contextualized embedding-based similarity metrics to evaluate summary quality, such as BERTScore~\cite{zhang2019bertscore}, MoverScore~\cite{zhao2019moverscore}, and Sentence Mover’s Similarity~\cite{clark2019sentence}. These metrics heavily depend on the pre-trained encoder to vectorize the candidate and ground truth summaries, potentially introducing inherent biases and suffering from low interpretability.

On the other hand, \citet{belz2008intrinsic} argue that this similarity reflects 'human-likeness' and may be unrelated to the final performance on generation tasks. BartScore~\cite{yuan2021bartscore} then conceptualizes the evaluation of generated text as a text generation problem. It argues that models trained to convert the generated text to/from a reference output or the source text will achieve higher scores when the generated text is better. Meanwhile, researchers are also exploring reference-free automatic evaluation metrics such as SummaQA~\cite{scialom2019answers}, BLANC~\cite{vasilyev2020fill}, and SUPERT~\cite{gao2020supert}.

\subsubsection{Factual Consistency}

The factual consistency of generated summaries is crucial for their real-world applications~\cite{maynez2020faithfulness,kryscinski-etal-2020-evaluating,zhang2022improving}. Researchers have delved into metrics to automatically evaluate summary faithfulness based on text entailment or question answering (QA).

The text entailment-based approach evaluates factual inconsistency by verifying the summary against the original document. FactCC~\cite{kryscinski-etal-2020-evaluating} is a weakly supervised BERT-based model metric that uses rule-based transformations applied to source document sentences. DAE~\cite{goyal2020evaluating} decomposes entailment at the level of dependency arcs, examining the semantic relationships within the generated output and input. SummaC~\cite{laban2022summac} segments documents into sentence units and aggregates scores between pairs of sentences.

QA-based metrics, on the other hand, employ a question-generation model to generate questions from a given summary and evaluate factual consistency by the extent to which the summary provides sufficient information to answer these questions. FEQA~\cite{durmus2020feqa} and QAGS~\cite{wang2020asking} measure summary factual consistency by determining how well a summary answers questions posed on its corresponding source document. Questeval~\cite{scialom2021questeval} unifies precision and recall-based QA metrics, measuring summary factual consistency by assessing how many generated questions can be answered by the summary.

\subsubsection{Coherence and Redundancy}

SNaC~\cite{goyal2022snac} is a narrative coherence evaluation framework rooted in fine-grained annotations, specifically designed for long summaries. \citet{peyrard2017learning} propose measuring the redundancy extent of a summary using its unique n-gram ratio, which is the percentage of unique n-grams. \citet{xiao2020systematically} introduce another metric to measure summary redundancy as the inverse of a diversity metric with length normalization, where diversity is defined as the entropy of unigrams in the document.

\subsection{Summarization Datasets}

The availability of public datasets for text summarization has been a major driving force behind the rapid advancement in recent years. These datasets vary across multiple dimensions, including domain, formats, size, and the number of gold summaries they provide. Here, we summarize the characteristics of these datasets and provide an overview in Table~\ref{tab:dataset}, which includes details such as the standard train-validation-test split, language, domain, summary format, and URL links. Our primary focus is on datasets containing summaries of texts written in English.

\subsubsection{Short News Summarization Dataset}\hfill

\textbf{CNN/DM}~\cite{hermann2015teaching} is the most widely-used dataset for summarization. It contains news articles and associated highlights as summaries written by journalists from CNN and the DailyMail. The original dataset is created for machine comprehension and adapted for summarization by \cite{nallapati2016abstractive}.

\textbf{XSum}~\cite{narayan2018don} is a dataset consisting of one-sentence summaries derived from online articles published by BBC. However, it has been noted that some summaries in the XSum dataset suffer from low faithfulness, and contain information that cannot be directly inferred from the source document~\cite{maynez2020faithfulness}.

\textbf{NYT}~\cite{sandhaus2008new} contains news articles with abstractive
summaries from the New York Times. Input documents are commonly truncated to $768$ tokens following~\cite{zhang2019hibert}.

\textbf{NEWSROOM}~\cite{grusky2018newsroom} is a large dataset containing article-summary pairs written by authors and
editors in the newsrooms of $38$ major publications between 1998 and 2017.

\textbf{Gigaword}~\cite{rush2015neural} contains around $4$ million headline-article pairs extracted from news articles (seven publishers) from the Gigaword corpus.

\textbf{CCSUM}~\cite{jiang2024ccsum} is a more recent news summary dataset that contains 1.3 million high-quality recent training news articles, filtered from CommonCrawl News articles.

\subsubsection{Other Domains}\hfill

\textbf{WikiHow}~\cite{koupaee2018wikihow} is a diverse dataset extracted from online knowledge base. Each article consists of a number of steps, and the gold-standard summary for each document is the concatenation of bold statements.

\textbf{Reddit}~\cite{kim2019abstractive} contains 120K posts of informal stories from the TIFU online discussion forum subreddit. The summaries are highly abstractive and informal.

\textbf{SAMSum}~\cite{gliwa2019samsum} consists of approximately 16,000 single-document, text message-like dialogues. These documents are created by linguists fluent in English and cover a broad spectrum of formality levels and discussion topics.

\textbf{AESLC}~\cite{zhang2019email} is an email domain dataset for the task of automatically generating an email subject line from the email body.

\subsubsection{Long Document Summarization Dataset}\hfill

\textbf{PubMed and arXiv}~\cite{cohan2018discourse} are two widely-used long document
datasets of scientific publications from arXiv.org and
PubMed. The task is to generate the abstract of the paper from
the body content.

\textbf{BIGPATENT}~\cite{sharma2019bigpatent} consists of 1.3 million US patents along with human summaries under nine patent
classification categories.

\textbf{BillSum}~\cite{kornilova2019billsum} contains 23k US Congressional and California state bills and human-written reference summaries from the 103rd-115th (1993-2018) sessions of Congress.

\textbf{FINDSum}~\cite{liu2023neural} is a new large-scale dataset for long text and multi-table summarization built on $21,125$ annual reports from $3,794$ companies.

\subsubsection{Multi-document Summarization Dataset}\hfill

\textbf{DUC}~\cite{over2007duc} presents an overview of the datasets provided by the Document Understanding Conferences (DUC). The DUC datasets from 2005 to 2007 include multi-document summarization datasets in the news domain, with 10-30 documents and 3-4 human-written summaries per cluster.

\textbf{MultiNews}~\cite{fabbri2019multi} is a large-scale multi-document news summarization dataset from the newser.com website with human-written summaries. Each summary is professionally written by editors, and the dataset covers diverse news sources.

\textbf{WikiSum}~\cite{liu2018generating} is a multi-document summarization dataset where each summary is a
Wikipedia article and the source documents are
either citations in the reference section or the web search results of section titles.

\textbf{WCEP}~\cite{ghalandari2020large} is
built based on news events from Wikipedia Current
Events Portal (WCEP) and similar articles in the CommonCrawl News dataset. It features high alignment with several real-world
industrial use cases and has $235$ articles per cluster on average.

\textbf{Multi-XScience}~\cite{lu2020multi} is a multi-document summarization dataset created from scientific articles. The summaries are paragraphs of the related work section, while source documents include the abstracts of the query and referred papers.

\textbf{Yelp}~\cite{chu2018unsupervised} is a customer review dataset created from the Yelp Dataset Challenge. The task is to summarize multiple reviews of a business.

\subsubsection{Query-focused Summarization Dataset}\hfill

\textbf{QMSum}~\cite{zhong2021qmsum} is a meeting summarization dataset that consists of 1,808 query-summary pairs over 232 meetings in multiple domains. The dataset also contains the main topics of each
meeting and the ranges of relevant text spans for
the annotated topics and each query.


\textbf{NewTS}~\cite{bahrainian2022newts} is an
aspect-focused summarization dataset derived from
the CNN/DM dataset and contains two summaries
focusing on different topics for the same news.

\textbf{TD-QFS}~\cite{baumel2016topic} is centered on medical texts and comprises short keyword queries across 4 clusters, each containing 185 documents. The clusters in TD-QFS exhibit lower topical concentration, featuring larger quantities of query-irrelevant information.

\subsubsection{Others}\hfill

\textbf{XL-Sum}~\cite{hasan2021xl} is a comprehensive and diverse dataset comprising 1 million professionally annotated article-summary pairs from BBC. The dataset covers 44 languages and is highly abstractive, concise, and of high quality.


\begin{table*}[!ht]
\centering
  \caption{Overview of Common Text Summarization Datasets.}
  \label{tab:dataset}
  \scalebox{0.90}{
\begin{tabular}{c|c|c|c|c|c|c}
\toprule
Dataset & Size & Language & Domain &Format &Source& Link \\

\midrule
CNN/DM& $287,084/13,367/11,489$ & English & News & SDS& \cite{hermann2015teaching}&\href{https://github.com/abisee/cnn-dailymail}{url}\\

XSum & $203,028/11,273/11,332$ & English & News & SDS& \cite{narayan2018don}&\href{https://github.com/EdinburghNLP/XSum}{url} \\

NYT & $137,778/17,222/7,223$ & English & News & SDS& \cite{sandhaus2008new}&\href{https://github.com/outerproduct/nyt-summ}{url} \\

NEWSROOM &$995,041/108,837/108,862$ & English &News &SDS& \cite{grusky2018newsroom}&\href{https://lil.nlp.cornell.edu/newsroom/index.html}{url} \\ 

Gigaword & $3,803,957/189,651/1,951$ & English & News&SDS&\cite{rush2015neural} &\href{https://github.com/harvardnlp/sent-summary}{url}\\

CCSUM & $1,349,911/10,000/10,000$ & English & News & SDS& \cite{jiang2024ccsum} &\href{https://github.com/amazon-science/ccsum}{url}\\
\hline

WikiHow & $ 168,126/6,000/6,000$ & English & Knowledge Base & SDS& \cite{koupaee2018wikihow} &\href{https://github.com/mahnazkoupaee/WikiHow-Dataset}{url}\\ 

Reddit & $41,675/645/645$ & English & Social Media & SDS& \cite{kim2019abstractive} &\href{https://github.com/ctr4si/MMN}{url}\\ 

SAMSum & $14,732/818/819$ & English & Dialogue& SDS& \cite{gliwa2019samsum} &\href{https://huggingface.co/datasets/samsum}{url}\\ 

MediaSum & $463,596$ & English & Dialogue& SDS& \cite{zhu2021mediasum} &\href{https://github.com/zcgzcgzcg1/MediaSum}{url}\\ 

AESLC & $14,436/1,960/1,906$ & English & Email & SDS& \cite{zhang2019email} &\href{https://github.com/ryanzhumich/AESLC}{url}\\

\hline
ArXiv & $201,427/6,431/6,436$ & English & Scientific Paper & SDS& \cite{cohan2018discourse}&\href{https://github.com/armancohan/long-summarization}{url} \\

PubMed & $112,291/6,402/6,449$ & English & Scientific Paper & SDS& \cite{cohan2018discourse}&\href{https://github.com/armancohan/long-summarization}{url} \\ 

BIGPATENT & $1,207,222/67,068/67,072$ & English &Patent&SDS& \cite{sharma2019bigpatent}&\href{https://evasharma.github.io/bigpatent/}{url}\\ 

BillSum & $18,949/1,237/3,269$ & English & Bill &SDS& \cite{kornilova2019billsum}&\href{https://github.com/FiscalNote/BillSum}{url}\\ 

FINDSum & $42,250$ & English & Report &SDS& \cite{liu2023neural}&\href{https://github.com/StevenLau6/FINDSum}{url}\\

\hline

DUC 05/06/07 & $50 \times 32/25/10$ & English & News & MDS& \cite{over2007duc} &\href{https://www-nlpir.nist.gov/projects/duc/data.html}{url}\\

MultiNews & $ 44,972/5,622/5,622$ & English & News & MDS& \cite{fabbri2019multi}&\href{https://github.com/Alex-Fabbri/Multi-News}{url}\\ 

WikiSum & 1.5m/38k/38k & English & Wikipedia & MDS& \cite{liu2018generating}&\href{https://registry.opendata.aws/wikisum/}{url} \\

WCEP & $8,158/1,020/1,022 $ & English & Wikipedia & MDS& \cite{ghalandari2020large}&\href{https://github.com/complementizer/wcep-mds-dataset}{url} \\ 

Multi-XScience & $30,369/5,066/5,093$ & English & Scientific Paper & MDS& \cite{lu2020multi} &\href{https://github.com/yaolu/Multi-XScience}{url}\\ 

Yelp & $1,038,184/129,856/129,840$ & English & Review & MDS& \cite{chu2018unsupervised}&\href{https://github.com/sosuperic/MeanSum}{url}\\ 

\hline

QMSum & $1,257/272/279$ & English & Meeting & QFS& \cite{zhong2021qmsum}&\href{https://github.com/Yale-LILY/QMSum}{url}\\ 

NewTS & $4800/-/1200$ & English & News & QFS& \cite{bahrainian2022newts} &\href{https://github.com/ali-bahrainian/NEWTS}{url}\\


TD-QFS & $3,400$ & English & Medical & QFS& \cite{baumel2016topic}&\href{https://talbaumel.github.io/TD-QFS/dataset.html}{url} \\

\hline

XL-Sum & $1,005,292$ & Multiple & News & SDS& \cite{hasan2021xl}&\href{https://github.com/csebuetnlp/xl-sum}{url}\\


\hline

\end{tabular}}
\end{table*}

\section{Summarization Methods Prior to LLMs}
\label{sec:prior}
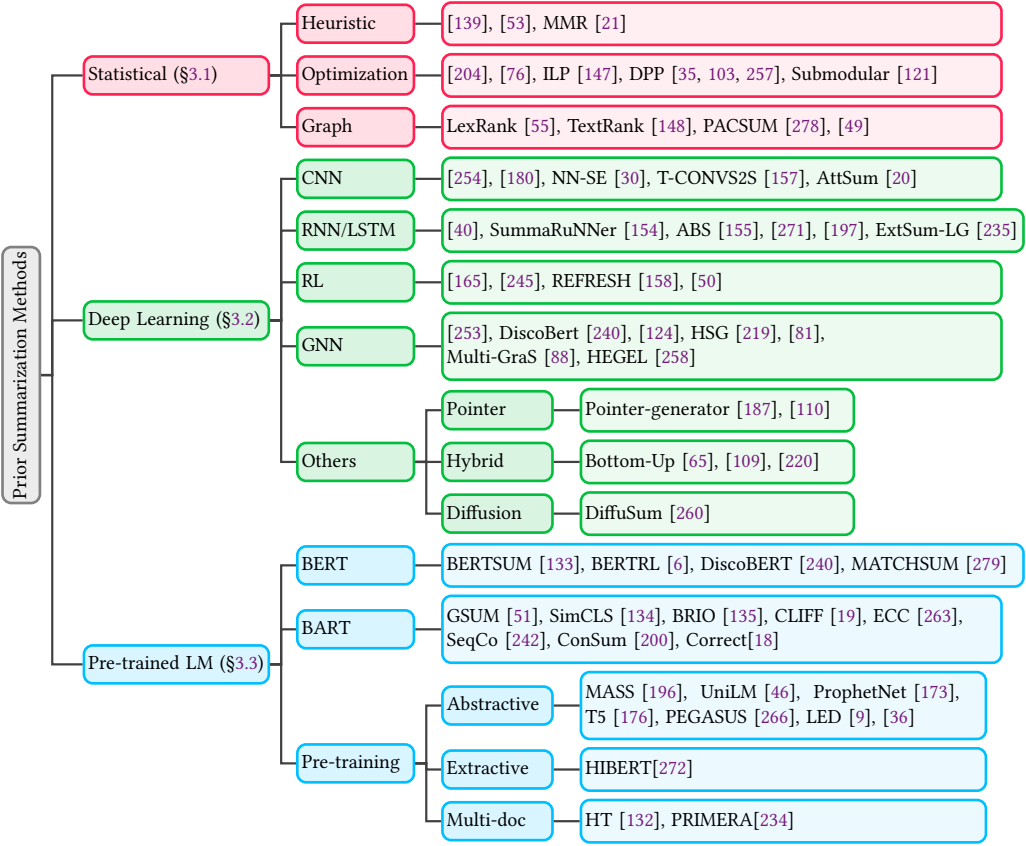
\begin{figure*}[tp]
    \centering
    \tikzstyle{my-box}=[
    rectangle,
    draw=hidden-draw,
    rounded corners,
    text opacity=1,
    minimum height=1.5em,
    minimum width=5em,
    inner sep=2pt,
    align=center,
    fill opacity=.5,
    ]
    \tikzstyle{leaf}=[my-box, minimum height=1.5em,
        fill=hidden-orange!60, text=black, align=left,font=\small,
        inner xsep=2pt,
        inner ysep=4pt,
    ]
    \resizebox{0.99\textwidth}{!}{
        \begin{forest}
            forked edges,
            for tree={
                grow=east,
                reversed=true,
                anchor=base west,
                parent anchor=east,
                child anchor=west,
                base=left,
                font=\small,
                rectangle,
                draw=hidden-draw,
                rounded corners,
                align=left,
                minimum width=4em,
                edge+={darkgray, line width=1pt},
                s sep=3pt,
                inner xsep=2pt,
                inner ysep=3pt,
                ver/.style={rotate=90, child anchor=north, parent anchor=south, anchor=center},
            },
            where level=1{text width=8em,font=\small,}{},
            where level=2{text width=4.9em,font=\small,}{},
            where level=3{text width=4.6em,font=\small,}{},
            where level=4{text width=6.8em,font=\small,}{},
            where level=5{text width=10.8em,font=\small,}{},
            [
                Prior Summarization Methods, draw=gray, color=gray!100, fill=gray!15, very thick, text=black, ver
                [
                   Statistical (\S \ref{sec:statistic}), color=awesome!100, fill=awesome!15, very thick, text=black
                    [
                        Heuristic, color=awesome!100, fill=awesome!15, very thick, text=black
                        [
                            \cite{luhn1958automatic}{, }\cite{edmundson1969new}{, }MMR~\cite{carbonell1998use}
                                    , leaf, text width=25em, color=awesome!100, fill=awesome!15, very thick, text=black
                        ]
                    ]
                    [
                        Optimization, color=awesome!100, fill=awesome!15, very thick, text=black
                        [
                            \cite{takamura2009text}{, }\cite{hirao2013single}{, }ILP~\cite{mcdonald2007study}{, }DPP~\cite{kulesza2012determinantal,cho2019improving,zhang2023unsupervised}{, }Submodular~\cite{lin2011class}
                                    , leaf, text width=25em, color=awesome!100, fill=awesome!15, very thick, text=black
                        ]
                    ]
                    [
                        Graph, color=awesome!100, fill=awesome!15, very thick, text=black
                        [
                            LexRank~\cite{erkan2004lexrank}{, }TextRank~\cite{mihalcea2004textrank}{, }PACSUM~\cite{zheng2019sentence}{, }\cite{dong2020discourse}
                                    , leaf, text width=25em, color=awesome!100, fill=awesome!15, very thick, text=black
                        ]
                    ]
                ]
                [
                    Deep Learning (\S \ref{sec:dl}), color=darkpastelgreen!100, fill=darkpastelgreen!15, very thick, text=black
                    [
                        CNN, color=darkpastelgreen!100, fill=darkpastelgreen!15, very thick, text=black
                        [
                            \cite{yin2015optimizing}{,}~\cite{rush2015neural}{,}~NN-SE~\cite{cheng2016neural}{,}~T-CONVS2S~\cite{narayan2018don}{, }AttSum~\cite{cao2016attsum}, leaf, text width=25em, color=darkpastelgreen!100, fill=darkpastelgreen!15, very thick, text=black
                        ]
                    ]
                    [
                        RNN/LSTM, color=darkpastelgreen!100, fill=darkpastelgreen!15, very thick, text=black
                        [
                            \cite{cohan2018discourse}{, }SummaRuNNer~\cite{nallapati2017summarunner}{, }ABS~\cite{nallapati2016abstractive}{, }\cite{zhang2018neural}{, }\cite{song2018structure}{, }ExtSum-LG~\cite{xiao2019extractive}, leaf, text width=26em, color=darkpastelgreen!100, fill=darkpastelgreen!15, very thick, text=black
                        ]
                    ]
                    [
                       RL, color=darkpastelgreen!100, fill=darkpastelgreen!15, very thick, text=black
                        [
                            \cite{paulus2018deep}{, }\cite{yadav2021reinforcement}{, }REFRESH~\cite{narayan2018ranking}{, }\cite{dong2018banditsum}, leaf, text width=25em, color=darkpastelgreen!100, fill=darkpastelgreen!15, very thick, text=black
                        ]
                    ]
                    [
                        GNN, color=darkpastelgreen!100, fill=darkpastelgreen!15, very thick, text=black
                        [
                            \cite{yasunaga2017graph}{, }DiscoBert~\cite{xu2019discourse}{, }\cite{liu2018toward}{, }HSG~\cite{wang2020heterogeneous}{, }\cite{huang2020knowledge}{, }\\Multi-GraS~\cite{jing2021multiplex}{, }HEGEL~\cite{zhang2022hegel}, leaf, text width=25em, color=darkpastelgreen!100, fill=darkpastelgreen!15, very thick, text=black
                        ]
                    ]
                    [
                        Others, color=darkpastelgreen!100, fill=darkpastelgreen!15, very thick, text=black
                        [
                            Pointer, color=darkpastelgreen!100, fill=darkpastelgreen!15, very thick, text=black
                            [
                            Pointer-generator~\cite{see2017get}{, }\cite{lebanoff2018adapting}, leaf, text width=12em, color=darkpastelgreen!100, fill=darkpastelgreen!15, very thick, text=black
                            ]
                        ]
                        [
                            Hybrid, color=darkpastelgreen!100, fill=darkpastelgreen!15, very thick, text=black
                            [
                            Bottom-Up~\cite{gehrmann2018bottom}{, }\cite{lebanoff2019scoring}{, }\cite{wang2022salience}, leaf, text width=12em, color=darkpastelgreen!100, fill=darkpastelgreen!15, very thick, text=black
                            ]
                        ]
                        [
                            Diffusion, color=darkpastelgreen!100, fill=darkpastelgreen!15, very thick, text=black
                            [
                            DiffuSum~\cite{zhang2023diffusum}, leaf, text width=12em, color=darkpastelgreen!100, fill=darkpastelgreen!15, very thick, text=black
                            ]
                        ]
                    ]
                ]
                [
                     Pre-trained LM (\S \ref{sec:pre-train}), color=capri!100, fill=capri!15, very thick, text=black
                    [
                        BERT, color=capri!100, fill=capri!15, very thick, text=black
                        [
                            BERTSUM~\cite{liu2019text}{, }BERTRL~\cite{bae2019summary}{, }DiscoBERT~\cite{xu2019discourse}{, }MATCHSUM~\cite{zhong2020extractive}
                                    , leaf, text width=26em, color=capri!100, fill=capri!15, very thick, text=black
                        ]
                    ] 
                    [
                        BART, color=capri!100, fill=capri!15, very thick, text=black
                        [
                            GSUM~\cite{dou2020gsum}{, }SimCLS~\cite{liu2021simcls}{, }BRIO~\cite{liu2022brio}{, }CLIFF~\cite{cao2021cliff}{, }ECC~\cite{zhang2022improving}{, }\\SeqCo~\cite{xu2022sequence}{, }ConSum~\cite{sun2021alleviating}{, }Correct\cite{cao2020factual}
                                    , leaf, text width=25em, color=capri!100, fill=capri!15, very thick, text=black
                        ]
                    ]
                    [
                        Pre-training, color=capri!100, fill=capri!15, very thick, text=black
                        [
                            Abstractive, color=capri!100, fill=capri!15, very thick, text=black
                            [
                                MASS~\cite{song2019mass}{, } UniLM~\cite{dong2019unified}{, } ProphetNet~\cite{qi2020prophetnet}{, }\\T5~\cite{raffel2020exploring}{, }PEGASUS~\cite{zhang2020pegasus}{, }LED~\cite{beltagy2020longformer}{, }\cite{cho2022toward}
                                    , leaf, text width=18em, color=capri!100, fill=capri!15, very thick, text=black
                            ]
                        ]
                        [
                            Extractive, color=capri!100, fill=capri!15, very thick, text=black
                            [
                                HIBERT\cite{zhang2019hibert}
                                    , leaf, text width=18em, color=capri!100, fill=capri!15, very thick, text=black
                            ]
                        ]
                        [
                            Multi-doc, color=capri!100, fill=capri!15, very thick, text=black
                            [
                                HT~\cite{liu2019hierarchical}{, }PRIMERA\cite{xiao2022primera}
                                    , leaf, text width=18em, color=capri!100, fill=capri!15, very thick, text=black
                            ]
                        ]
                    ]
                ]
            ]
        \end{forest}
}
    \caption{Taxonomy of Representative Summarization Methods prior to LLMs.}
    \label{fig:pre_tax}
\end{figure*}

Before delving into the new opportunities enabled by recent advancements in LLMs, this section reviews the significant progress previously achieved in summarization by investigating representative summarization methods and algorithms, as outlined in the overview taxonomy shown in Fig.~\ref{fig:pre_tax}. This comprehensive discussion encompasses the three major summarization paradigms prior to LLMs: statistical methods (\S~\ref{sec:statistic}), deep learning-based approaches (\S~\ref{sec:dl}), and PLM-based fine-tuning approaches (\S~\ref{sec:pre-train}). The details, including method category, backbone, and datasets of these representative methods, are summarized in Table~\ref{tab:method}.

\subsection{Statistical Summarization Methods} 
\label{sec:statistic}

In the early stages of text summarization systems, research primarily focuses on extractive summarization and relies on traditional statistical approaches to automatically create summaries. These systems mostly use frequency-based features such as TF-IDF and hand-crafted features~\cite{edmundson1969new} to model text data. Representative methods from this stage include heuristic-based methods, optimization-based methods, and graph-based ranking methods.

\subsubsection{Heuristic-based Methods}

One of the earliest summarization approaches is by \citet{luhn1958automatic}. This method hypothesizes that the importance of sentences can be measured by the frequency of specific content words (keywords). It scores sentences based on importance and then extracts the high-scoring ones to generate literature abstracts.

\citet{edmundson1969new} later proposes modeling sentences based on a combination of handcrafted features, including word frequency within the article, sentence position, the number of words in the article title or section headings, and the frequency of cue words (keywords). A simple linear summation is then applied to score and rank the sentences.

Maximal Marginal Relevance (MMR)~\cite{carbonell1998use} strives to reduce redundancy while maintaining query relevance in selecting sentences for extractive summarization. It employs a greedy approach to combine sentence relevance with information novelty.

\subsubsection{Optimization-based Methods}

The sentence selection problem in extractive summarization has been formulated as a constrained optimization problem to obtain a globally optimal set of sentences. Summary sentences are selected to maximize the coverage of important source content while minimizing redundancy and adhering to a length constraint. Text summarization has been formalized as various optimization problems, such as the maximum coverage problem~\cite{takamura2009text} and the tree knapsack problem~\cite{hirao2013single}, and addressed with optimization methods including integer linear programming (ILP)~\cite{mcdonald2007study}, determinantal point process (DPP)~\cite{kulesza2012determinantal, cho2019improving}, and submodular functions~\cite{lin2011class}.

\subsubsection{Graph-based Methods} 

LexRank~\cite{erkan2004lexrank} and TextRank~\cite{mihalcea2004textrank} are two seminal works that introduce the concept of representing documents as graphs and formulating extractive summarization as identifying the most central nodes in a graph, inspired by the PageRank algorithm \cite{brin1998anatomy}. The underlying premise is that the centrality of a node in the graph can serve as an approximation of the significance of its corresponding sentence within the document.

\begin{figure}[!tp]
    \centering
    \includegraphics[width=0.9\textwidth]{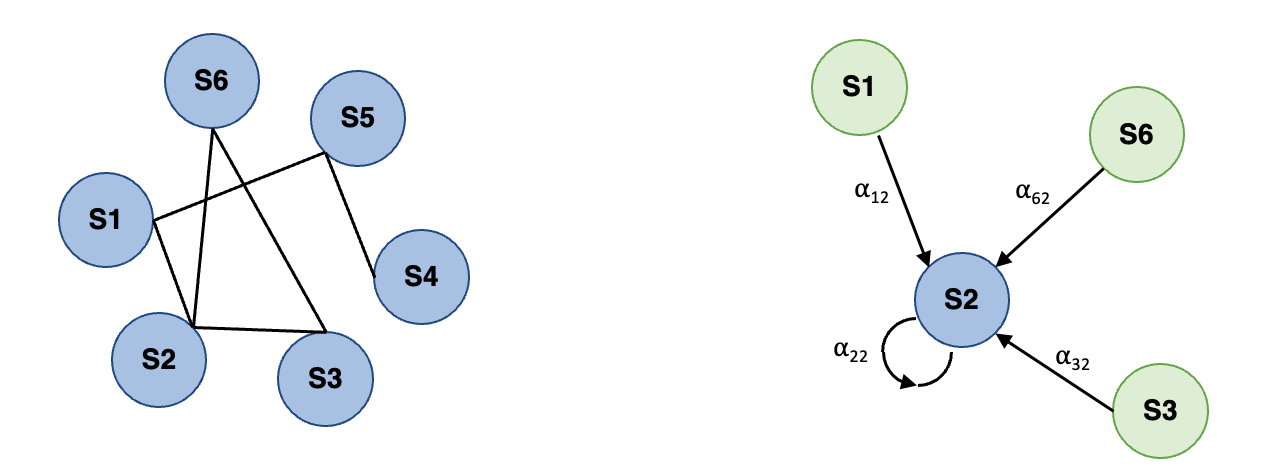}
    \caption{\textbf{Left}: Illustration of representing a document consisting of 6 sentences $\{s1, ..., s6\}$ as a graph. Each node represents one sentence and the edge weights represent sentence similarities. \textbf{Right}: Illustration of graph attention by node 2 on its neighborhood. Here $\alpha_{ij}$ denotes
the normalized attention scores between node i and node j. 
    }
    \label{fig:graph}
\end{figure}

In the left of Figure~\ref{fig:graph}, we illustrate
the graph representation of a document. To construct the graph $G=(V,E)$ from a document, each node $v\in V$ represents a sentence in the document, and each edge $e\in E$ represents the similarity between connected pairs of nodes. Edges with weights lower than a
pre-defined threshold are pruned prior to graph computation to avoid a fully connected graph. Subsequently, edge weights are normalized, and the Markov chain is iteratively applied to the graph to determine node centrality. Finally, sentences are ranked and selected based on their centrality for inclusion in the summary.

More recently, PACSUM~\cite{zheng2019sentence} revisited the TextRank algorithm and proposed constructing graphs with directed edges. They argue that the relative position of nodes in a document influences the contribution of any two nodes to their centrality. Additionally, PACSUM integrates neural representations to improve sentence similarity computation.

\subsection{Deep Learning Summarization Methods}
\label{sec:dl}
Advances in deep neural networks and the availability of large-scale text corpora have significantly accelerated progress in various NLP tasks. The development of word embedding techniques, such as Word2Vec~\cite{mikolov2013efficient} and GloVe~\cite{pennington2014glove}, has enabled deep learning models to represent text data as sequences of tokens, automatically learning continuous sentence and document representations from data and eliminating the need for human-engineered features. Common deep learning frameworks used in NLP include convolutional neural networks (CNNs), recurrent neural networks (RNNs), long short-term memory (LSTM) networks~\cite{hochreiter1997long}, reinforcement learning (RL), and graph neural networks (GNNs). Summarization research in the deep learning stage still predominantly focuses on extractive summarization in a supervised manner. However, abstractive summarization methods are beginning to emerge, leveraging frameworks such as sequence-to-sequence (seq2seq) models~\cite{sutskever2014sequence}, pointer networks~\cite{vinyals2015pointer}, and attention mechanisms.

\subsubsection{CNN-based Methods}

\begin{figure}[!htbp]
    \centering
    \includegraphics[width=0.95\textwidth]{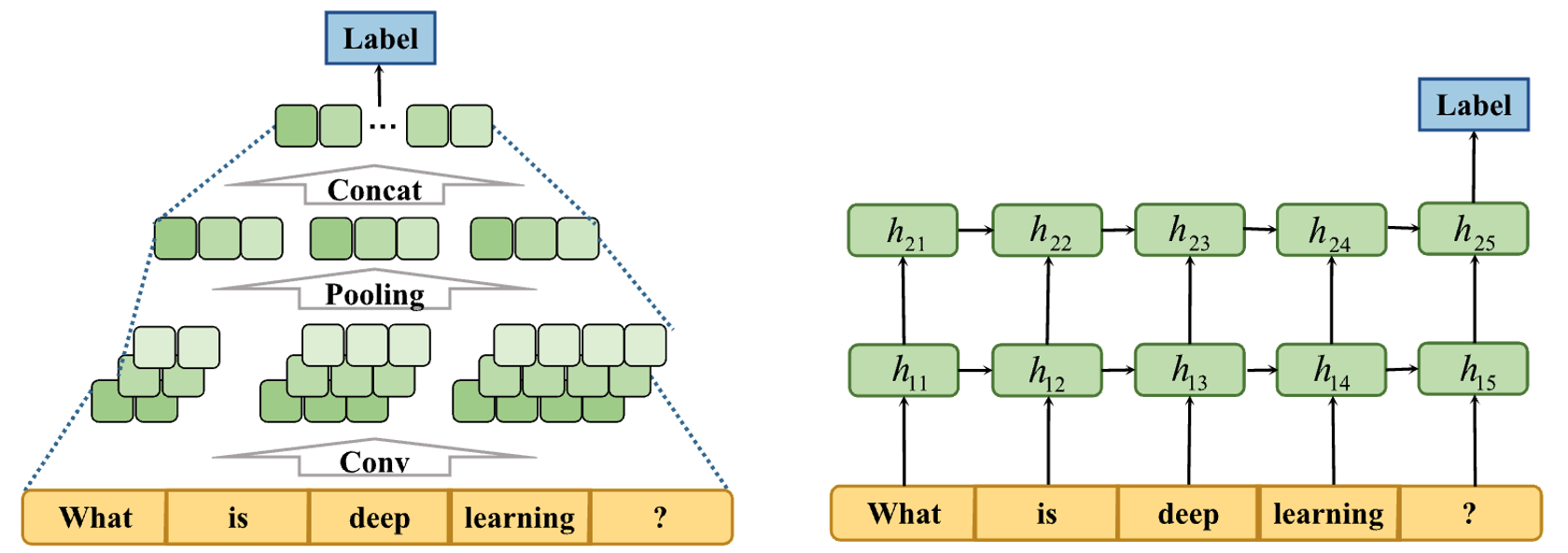}
    \caption{Illustration of the CNN model architecture (left) and the RNN model architecture (right) (taken from~\cite{li2022survey}).
    }
    \label{fig:cnnrnn}
\end{figure}

After the successful application of CNNs for learning continuous document representations in text classification tasks \cite{kim2014convolutional}, \citet{yin2015optimizing} propose a CNN-based neural network language model. This model projects sentences into dense distributed representations and employs a diversified selection process to extract fewer redundant sentences for summaries. \citet{rush2015neural} apply CNN for abstractive summarization with the sequence-to-sequence model structure. It uses CNN to encode the source and a context-sensitive attentional feed-forward neural network to generate the summary.

NN-SE~\cite{cheng2016neural} introduces a hierarchical document encoder that combines a convolutional sentence encoder for learning sentence representations with a recurrent document encoder for learning document representations. Additionally, NN-SE incorporates an attention-based neural sentence extractor to select sentences or words for summarization.

T-CONVS2S~\cite{narayan2018don} proposes a topic-aware convolutional sequence-to-sequence learning approach for the extreme summarization task. The distributions obtained from Latent Dirichlet Allocation (LDA)~\cite{blei2003latent} are used as additional input. The convolutional encoder associates each word with a topic vector, while the convolutional decoder conditions each word prediction on a document topic vector.

For query-focused summarization, AttSum~\cite{cao2016attsum} proposes a method that jointly ranks query relevance and sentence saliency. It automatically learns distributed representations for sentences and queries within the latent space using CNNs.

\subsubsection{RNN/LSTM-based Methods}
SummaRuNNer~\cite{nallapati2017summarunner} is a representative RNN-based extractive summarization model. It employs a two-layer bidirectional RNN, with the bottom layer operating at the word level within each sentence and the top layer running over sentences. An accumulated document representation is generated through a weighted sum of sentence representations, and a logistic classifier is used to make predictions.

ABS~\cite{nallapati2016abstractive} first applies the attentional encoder-decoder RNN for abstractive summarization, proposing to model rare or unseen words with a switching generator-pointer mechanism and to capture the document structure with hierarchical attention.

For long document summarization, \citet{cohan2018discourse} propose a pioneering model for abstractive summarization of long documents. Their approach employs a hierarchical encoder to capture the discourse structure (scientific paper sections) of a document and an attentive discourse-aware decoder to produce the summary. Building on this, ExtSum-LG~\cite{xiao2019extractive} proposes using the global context of the entire document along with the local context. It applies the LSTM-Minus method to capture the local context of each sentence.

\subsubsection{RL-based Methods}


Reinforcement learning is a popular alternative for training abstractive and extractive summarization systems, addressing the exposure bias and objective mismatch issues associated with the negative log-likelihood (NLL) training loss~\cite{fu2022inverse}. 

A seminal work in this domain is REFRESH~\cite{narayan2018ranking}, which applies RL to extractive summarization by conceptualizing it as a sentence ranking task. This study introduces an objective function that integrates the maximum-likelihood cross-entropy loss with rewards derived from policy gradient reinforcement learning. \citet{paulus2018deep} further advance this direction by introducing a neural network model with a novel intra-attention mechanism that attends to the input and the continuously generated output separately. They also propose a new training method that combines standard supervised word prediction with reinforcement learning. Moreover, \citet{yadav2021reinforcement} propose two novel rewards obtained from the downstream tasks of question-type identification and question-focus recognition to regularize the generation model.

\subsubsection{GNN-based Methods}

Researchers have also explored the use of Graph Neural Networks (GNNs)~\cite{velivckovic2017graph,kipf2016semi,zhang2020graph} for NLP due to their ability to better capture the structural information of a document, as well as their scalability and interpretability~\cite{yao2019graph,zhang2020text,zhang2023contrastive,salchner2022survey,zhao2024hierarchical}. GNNs typically use the adjacency matrix and initial node representation matrix as inputs, performing information aggregation on the graph through either spatial or spectral convolution. As illustrated in the right of Fig.~\ref{fig:graph}, Graph Attention Networks (GATs)~\cite{velivckovic2017graph} is one of the most popular GNNs that introduces attention mechanism to the
spatial convolution to filter unimportant neighbors. Here $\alpha_{ij}$ denotes
the normalized attention scores between node i and node j.


\citet{yasunaga2017graph} is one of the pioneering endeavors in applying GNN to extractive summarization tasks. Their approach involves modeling documents as approximate discourse graphs and leveraging Graph Convolutional Networks (GCN)~\cite{kipf2016semi} to encode sentence representations, which are subsequently used for predicting sentence salience. Similarly, DiscoBert~\cite{xu2019discourse} introduces a method that employs GCN on structural discourse graphs constructed from Rhetorical Structure Theory (RST) trees and coreference mentions. Moreover, HSG~\cite{wang2020heterogeneous} proposes an approach that views documents as heterogeneous graphs, incorporating semantic nodes of various granularity levels (words, sentences, documents). The model constructs word-document heterogeneous graphs and employs word nodes as the intermediary between sentences, enabling a more nuanced representation of document semantics.

Researchers have also explored various graph variants to enhance document modeling. Multi-GraS~\cite{jing2021multiplex} introduces the utilization of multiplex graphs to capture multiple types of inter-sentential relationships, such as semantic similarity and natural connection, while also modeling intra-sentential relationships. Furthermore, HEGEL~\cite{zhang2022hegel} proposes a novel approach by representing documents as hypergraphs to capture high-order cross-sentence relations instead of pairwise relations. This method integrates diverse types of sentence dependencies, including section structure, latent topics, and keyword coreference, thereby offering a comprehensive view of document semantics.

\subsubsection{Others}
In addition to the aforementioned models, here we introduce some representative individual models. 

\citet{see2017get} introduce the pointer-generator network~\cite{vinyals2015pointer} to mitigate the issue of inaccurate reproduction of factual details by seq2seq models. This approach enables the model to both generate words from the vocabulary using a generator and selectively copy content from the source using a pointer. \citet{lebanoff2018adapting} further adapt the model to the multi-document summarization setting with MMR.

Bottom-Up~\cite{gehrmann2018bottom} introduces a data-efficient content selector to identify potential output words in a source document. This selector is integrated as a bottom-up attention mechanism within the pointer-generator network, representing a hybrid approach that combines extractive and abstractive methods. Later, \citet{lebanoff2019scoring} propose selecting and fusing sentences into summaries.

DiffuSum~\cite{zhang2023diffusum} proposes a method to extract summaries by directly generating desired summary sentence representations using diffusion models and extracting sentences based on matching these representations. The approach optimizes a sentence encoder with a matching loss for aligning sentence representations and a multi-class contrastive loss for promoting representation diversity.

\subsection{Pre-trained Language Model Summarization Methods}
\label{sec:pre-train}

\begin{figure}[!htbp]
    \centering
\includegraphics[width=0.9\textwidth]{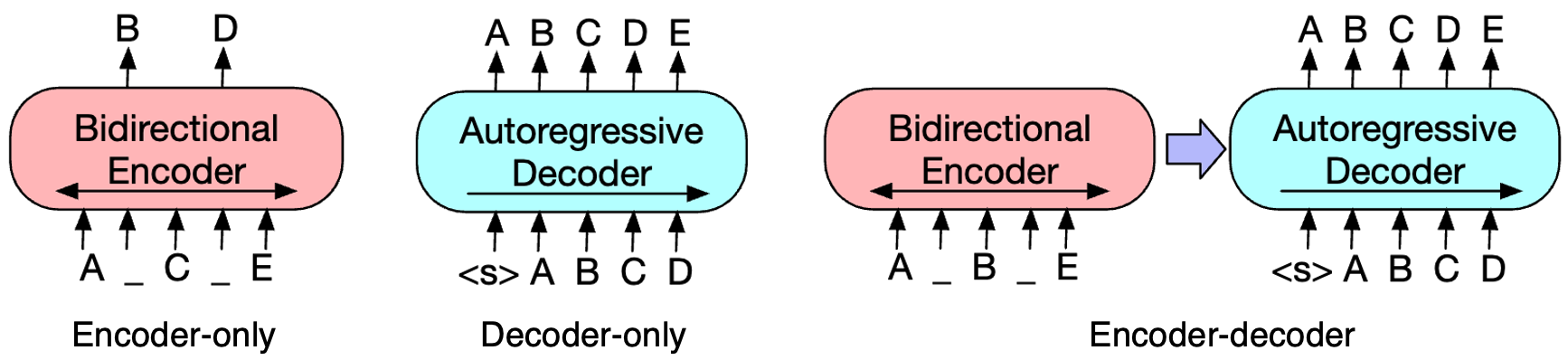}
    \caption{An illustration of the existing prevalent pre-training frameworks (taken from~\cite{lewis2019bart}). Encoder-only models mask random tokens during pre-training and encode documents bidirectionally. Decoder-only models predict tokens autoregressive conditioned on leftward context. Encoder-decoder models encode corrupted documents with a bidirectional model and then calculate generation likelihood with an autoregressive decoder.
    }
    \label{fig:encoder}
\end{figure}

With the advent of Transformer-based architectures \cite{vaswani2017attention} and the powerful large-scale self-supervised pre-training, summarization systems have experienced a substantial leap in performance over the past few years. The self-attention mechanism within the Transformer architecture facilitates parallel computation and enhances learning efficiency, allowing for effective training on large-scale unlabeled corpora. PLMs can learn universal language representations through this unsupervised training, offering superior model initialization for downstream tasks and eliminating the need to learn each new task from scratch. In this stage, summarization systems typically adapt PLM checkpoints, which have been trained in a self-supervised manner on large-scale corpora, and subsequently fine-tune these PLMs on domain- or task-specific training datasets.

As illustrated in Fig~\ref{fig:encoder}, three prevalent pre-training architectures are commonly used: encoder-only, encoder-decoder, and decoder-only. All these architectures employ the same self-attention layers as in vanilla Transformers to encode word tokens. The Bidirectional Encoder Representations from Transformers (BERT) \cite{devlin2018bert} marks the beginning of this phase, being the first widely adopted PLM that can be used for both extractive and abstractive summarization with its encoder-only architecture \cite{liu2019text}. Summarization research with PLMs has increasingly focused on abstractive summarization using encoder-decoder architecture models like T5 \cite{raffel2020exploring} and BART \cite{lewis2019bart} as the backbone. OpenAI has introduced decoder-only models with generative pre-training~\cite{radford2018improving}. Additionally, researchers have developed summarization-specific PLMs like PEGASUS \cite{zhang2020pegasus}, long-context PLMs like LED \cite{beltagy2020longformer}, and multi-document summarization PLMs like PRIMERA \cite{xiao2022primera}. 

In this discussion, we use BERT as a representative for encoder-only PLMs and BART for encoder-decoder models to explore their applications in text summarization in Sections~\ref{subsec:bert} and~\ref{subsec:bart}. We also examine representative efforts in pre-training from scratch in Section~\ref{subsec:pretrain}.

\subsubsection{BERT-based Methods}
\label{subsec:bert}
Encoder-only PLMs like BERT are predominantly used to convert natural language into continuous contextualized representations (embeddings), which can be further manipulated for downstream tasks. BERT is self-supervisedly pre-trained using the next sentence prediction (NSP) and masked language modeling (MLM) objectives on a large-scale unlabeled corpus. This encoder-only architecture makes BERT well-suited for various predictive modeling tasks, such as text classification and extractive summarization. Other popular encoder-only PLMs include RoBERTa \cite{liu2019roberta}, DistilBERT \cite{sanh2019distilbert}, ALBERT \cite{lan2019albert}, and Sentence-BERT \cite{reimers2019sentence}.

PreSum~\cite{liu2019text} stands as a seminal paper and the most popular baseline model for PLM-based summarization. It showcases how BERT could be effectively applied to text summarization and proposes a comprehensive framework for both extractive and abstractive summarization. They use several inter-sentence Transformer layers for extractive summarization and introduce a novel fine-tuning schedule that employs different optimizers for the encoder and the decoder, aiming to address the mismatch between these two components. Building upon PreSum, \citet{bae2019summary} further enhance it with reinforcement learning techniques to achieve summary-level scoring. They propose bridging extractive and rewrite models using policy-based reinforcement learning.

DiscoBERT~\cite{xu2019discourse} introduces an enhancement to BERT's document encoder by incorporating a graph encoder focused on structural discourse graphs. This augmentation aims to capture long-range contextual dependencies within the document.

MATCHSUM~\cite{zhong2020extractive} introduces a novel approach to extractive summarization by framing the task as a semantic text-matching problem. Instead of extracting sentences individually, this method matches a source document with candidate summaries in a semantic space. The model uses a Siamese-BERT architecture, comprising two BERT models with tied weights, alongside a cosine-similarity layer during the inference phase.

\subsubsection{BART-based Methods}
\label{subsec:bart}
Encoder-decoder PLMs like BART are widely used for NLP tasks that involve understanding input sequences and generating output sequences due to their ability to capture the mapping between these sequences. BART is self-supervisedly pre-trained by denoising and reconstructing text that has been corrupted with an arbitrary noising function on a large-scale unlabeled corpus. The encoder-decoder architecture makes BART well-suited for various sequence-to-sequence generation tasks, such as machine translation and abstractive summarization. Other popular encoder-decoder PLMs include MASS~\cite{song2019mass}, T5 \cite{raffel2020exploring}, UL2 \cite{tay2022ul2}, and instruction-tuned versions such as Flan-T5 and Flan-UL2 \cite{chung2024scaling}.

Besides directly applying PLMs for abstractive summarization, \citet{aghajanyan2020better} propose regularized fine-tuning rooted in trust-region theory to alleviate representational collapse, which is the degradation of generalizable representations
of pre-trained models during the fine-tuning stage. GSUM~\cite{dou2020gsum} introduces a guided summarization framework that incorporates external guidance as additional input to improve the faithfulness and controllability of the generated summaries.

Additionally, several research efforts have also explored the use of contrastive learning in fine-tuning BART. SimCLS~\cite{liu2021simcls} introduces a two-stage framework where the BART model generates candidate summaries, and a RoBERTa model, fine-tuned as an evaluation model, scores and selects the best candidate summaries, leveraging contrastive learning. SeqCo~\cite{xu2022sequence} and ConSum~\cite{sun2021alleviating} propose leveraging contrastive learning to improve BART fine-tuning, specifically to alleviate exposure bias. BRIO~\cite{liu2022brio} introduces a novel training paradigm that assumes a non-deterministic distribution, assigning probability mass to different candidate summaries based on their quality.

On the other hand, research on adapting the BART model has begun to focus on fine-tuning techniques to improve the faithfulness of generated summaries. These techniques include error correction \cite{cao2020factual}, contrastive training as implemented in CLIFF \cite{cao2021cliff}, post-processing methods \cite{chen2021improving}, knowledge graph methods~\cite{mao2022fact,zhu2020enhancing}, data filtering approaches \cite{nan2021entity}, and further advancements in controlled generation as seen in ECC \cite{zhang2022improving} and Spancopy~\cite{xiao2022entity}.

\subsubsection{Pre-training Methods}
\label{subsec:pretrain}

Besides the efforts in fine-tuning pre-trained language models on task-specific training data, researchers have also attempted to pre-train Transformer models with self-supervised objectives from scratch to better align with downstream tasks.

For extractive summarization, HIBERT~\cite{zhang2019hibert} proposes a hierarchical BERT model for document encoding, pre-trained using document masking and sentence prediction with unlabeled data. It leverages the hierarchical structure of documents to improve the quality of the extracted summaries.

Early efforts in pre-training language models for abstractive summarization include masked seq2seq pre-training in MASS~\cite{song2019mass}, unified pre-training in UniLM~\cite{dong2019unified}, and n-gram prediction pre-training in ProphetNet~\cite{qi2020prophetnet}. PEGASUS~\cite{zhang2020pegasus} is a popular PLM specifically designed for summarization. In PEGASUS, important sentences are removed or masked from an input document and are generated together as one output sequence from the remaining sentences, mimicking the process of creating an extractive summary.

Longformer-Encoder-Decoder (LED)~\cite{beltagy2020longformer} is a PLM specially designed for processing long documents. Longformer replaces the standard self-attention mechanism in Transformers with a combination of local windowed attention and task-motivated global attention. This modification allows the model to scale linearly with sequence length, enabling it to process documents consisting of thousands of tokens more efficiently. \citet{cho2022toward} later extend LED with joint training with segmentation and \citet{liu2021hetformer} improve the sparse attention with heterogeneous graphs.

For multi-document summarization, HT~\cite{liu2019hierarchical} augments vanilla Transformers with hierarchical inter-paragraph attention and document graph representation. PRIMERA~\cite{xiao2022primera} proposes the first PLM specifically designed for multi-document representation, based on the LED architecture. It introduces a new gap sentence generation pre-training objective tailored for the multi-document setting, teaching the model to connect and aggregate information across documents using entity pyramid masking.

\begin{table*}[!ht]
\centering
  \caption{Overview of Representative Summarization Methods before LLMs. Unsup. stands for unsupervised, Ext. stands for extractive, Abs. stands for abstractive.}
  \label{tab:method}
  \scalebox{0.85}{
  \begin{tabular}{c|c|c|c|c|l}
    \toprule
Paradigm&Category&Backbone&Year&Method&Datasets\\
    \midrule
\multirow{7}{*}{Statistical}&UnSup. Ext.&Heuristic&1998&MMR~\cite{carbonell1998use}&TIPSTER topic\\
\cline{2-6}

&UnSup. Ext.&Graph&2004&TextRank~\cite{mihalcea2004textrank}&Inspec\\
\cline{2-6}

&UnSup. Ext.&Graph&2004&LexRank~\cite{erkan2004lexrank}&DUC 03/04\\
\cline{2-6}

&UnSup. Ext.&Optimization&2007&ILP~\cite{mcdonald2007study}&DUC 02/05\\
\cline{2-6}

&UnSup. Ext.&Optimization&2011&Submodular~\cite{lin2011class}&DUC 03-07\\
\cline{2-6}

&UnSup. Ext.&Optimization&2012&DPP~\cite{kulesza2012determinantal}&DUC 03/04\\
\cline{2-6}

&UnSup. Ext.&Graph&2019&PACSUM~\cite{zheng2019sentence}&CNNDM, NTY, TTNEWS\\

\hline
\multirow{12}{*}{DL}

&Sup. Ext.&CNN,LSTM&2016&NN-SE ~\cite{cheng2016neural}&CNNDM, DUC2002\\
\cline{2-6}

&Sup. Ext.&RNN&2017&SummaRuNNer ~\cite{nallapati2017summarunner}&CNNDM, DUC2002\\
\cline{2-6}

&Sup. Abs.&Pointer network&2017&PTGEN ~\cite{see2017get}&CNNDM\\
\cline{2-6}

&Sup. Ext.&GNN&2017&GCN ~\cite{yasunaga2017graph}&DUC\\
\cline{2-6}

&Sup. Abs.&CNN&2018&T-CONVS2S ~\cite{narayan2018don}&CNNDM, XSUM\\
\cline{2-6}

&Sup. Hybrid&LSTM&2018&BottomUp ~\cite{gehrmann2018bottom}&CNNDM, NYT\\
\cline{2-6}

&Sup. Ext.&GRU&2018&NEUSUM ~\cite{zhou2018neural}&CNNDM\\
\cline{2-6}

&RL Ext.&RL&2018&REFRESH ~\cite{narayan2018ranking}&CNNDM\\
\cline{2-6}

&RL Ext.&RL&2018&Intra-attention ~\cite{paulus2018deep}&CNNDM, NYT\\

\cline{2-6}
&Sup. Ext.&LSTM&2019&ExtSum-LG~\cite{xiao2019extractive}&PubMed, ArXiv\\
\cline{2-6}

&Sup. Ext.&GNN&2020&HSG ~\cite{wang2020heterogeneous}&CNNDM, NYT, MultiNews\\
\cline{2-6}

&Sup. Ext.&GNN&2022&HEGEL ~\cite{zhang2022hegel}&arXiv, PubMed\\
\cline{2-6}

&Sup. Ext.&Diffusion&2023&DiffuSum~\cite{zhang2023diffusum}&CNNDM, XSUM, PubMed\\
\cline{2-6}

\hline
\multirow{18}{*}{PLM}

&Sup. Abs.&Transformer&2019&HT ~\cite{liu2019hierarchical}&WikiSum\\
\cline{2-6}

&Sup. Ext.&Transformer&2019&HIBERT ~\cite{zhang2019hibert}&CNNDM, NYT\\\cline{2-6}

&Sup. Abs.&Transformer&2019&MASS ~\cite{song2019mass}&CNNDM, XSUM, Gigaword\\
\cline{2-6}

&Sup. Ext.&BERT&2019&BERTSUMEXT~\cite{liu2019text}&CNNDM, NTY, XSUM\\
\cline{2-6}

&Sup. Abs.&BERT&2019&BERTSUMABS ~\cite{liu2019text}&CNNDM, NTY, XSUM\\
\cline{2-6}

&Sup. Abs.&Transformer&2019&UniLM~\cite{dong2019unified}&CNNDM, Gigaword\\
\cline{2-6}

&Sup. Abs.&Transformer&2019&BART ~\cite{lewis2019bart}&CNNDM, XSUM\\
\cline{2-6}

&Sup. Abs.&Transformer&2019&T5 ~\cite{raffel2020exploring}&CNNDM\\
\cline{2-6}

&Sup. Ext.&Graph, BERT&2020&DiscoBERT ~\cite{xu2019discourse}&CNNDM, NYT\\
\cline{2-6}

&Sup. Ext.&BERT&2020&MATCHSUM~\cite{zhong2020extractive}&CNNDM, XSUM, etc.\\
\cline{2-6}

&Sup. Abs.&Transformer&2020&PEGASUS ~\cite{zhang2020pegasus}&CNNDM, XSUM, etc.\\
\cline{2-6}

&Sup. Abs.&Transformer&2020&ProphetNet~\cite{qi2020prophetnet}&CNNDM, Gigaword\\
\cline{2-6}

&Sup. Abs.&Transformer&2020&LED ~\cite{beltagy2020longformer}&arXiv\\
\cline{2-6}

&Sup. Abs.&BART&2021&SimCLS ~\cite{liu2021simcls}&CNNDM, XSUM\\
\cline{2-6}

&Sup. Hybrid&BERT, BART&2021&GSUM~\cite{dou2020gsum}&CNNDM, XSUM, etc.\\
\cline{2-6}

&Sup. Abs.&BART&2021&CLIFF ~\cite{cao2021cliff}&CNN/DM, XSUM\\
\cline{2-6}

&Sup. Abs.&BART&2022&BRIO ~\cite{liu2022brio}&CNNDM, XSUM, NYT\\
\cline{2-6}

&Sup. Abs.&LED&2022&PRIMERA ~\cite{xiao2022primera}&WikiSum, arXiv, etc.\\

  \bottomrule
\end{tabular}}
\end{table*}
\section{LLM-based Summarization Research}
\label{sec:llm}

Recently, the advent of large language models like GPT-X has revolutionized the landscape of NLP and significantly influence the direction of research in text summarization. The majority of current summarization studies now employ LLMs as backbone models, leveraging zero-shot or few-shot settings instead of the previous fine-tuning approach. Given LLMs' strong capabilities in generating coherent and contextually relevant text, research is now primarily focused on abstractive summarization. This section aims to present a comprehensive overview of the recent research efforts in LLM-based text summarization, delineating the advancements and providing insights into the evolving research trends.

\begin{figure*}[tp]
    \centering
    \tikzstyle{my-box}=[
    rectangle,
    draw=hidden-draw,
    rounded corners,
    text opacity=1,
    minimum height=1.5em,
    minimum width=5em,
    inner sep=2pt,
    align=center,
    fill opacity=.5,
    ]
    \tikzstyle{leaf}=[my-box, minimum height=1.5em,
        fill=hidden-orange!60, text=black, align=left,font=\scriptsize,
        inner xsep=2pt,
        inner ysep=4pt,
    ]
    \resizebox{0.99\textwidth}{!}{
        \begin{forest}
            forked edges,
            for tree={
                grow=east,
                reversed=true,
                anchor=base west,
                parent anchor=east,
                child anchor=west,
                base=left,
                font=\small,
                rectangle,
                draw=hidden-draw,
                rounded corners,
                align=left,
                minimum width=4em,
                edge+={darkgray, line width=1pt},
                s sep=3pt,
                inner xsep=2pt,
                inner ysep=3pt,
                ver/.style={rotate=90, child anchor=north, parent anchor=south, anchor=center},
            },
            where level=1{text width=6em,font=\scriptsize,}{},
            where level=2{text width=4.9em,font=\scriptsize,}{},
            where level=3{text width=4.6em,font=\scriptsize,}{},
            where level=4{text width=6.8em,font=\scriptsize,}{},
            where level=5{text width=10.8em,font=\scriptsize,}{},
            [
                Summarization with LLMs, draw=gray, color=gray!100, fill=gray!15, very thick, text=black, ver
                [
                   Benchmarking (\S \ref{sec:benchmark}), color=lightcoral!100, fill=lightcoral!15, very thick, text=black
                    [
                        Generic, color=lightcoral!100, fill=lightcoral!15, very thick, text=black
                        [
                            \cite{goyal2022news}{, }\cite{zhang2024benchmarking}{, }\cite{pu2023summarization}{, }\cite{zhang2023extractive}{, }\cite{laskar2023building}{, }\cite{fu2024tiny}{, }\cite{wang2023cross}{, }\cite{huang2023embrace}
                                    , leaf, text width=14em, color=lightcoral!100, fill=lightcoral!15, very thick, text=black
                        ]
                    ]
                    [
                        Controllable, color=lightcoral!100, fill=lightcoral!15, very thick, text=black
                        [
                            \cite{yang2023exploring}{, }\cite{pu2023chatgpt}{, }\cite{liu2023benchmarking}
                                    , leaf, text width=14em, color=lightcoral!100, fill=lightcoral!15, very thick, text=black
                        ]
                    ]
                    [
                        Characteristics, color=lightcoral!100, fill=lightcoral!15, very thick, text=black
                        [
                            Factuality, color=lightcoral!100, fill=lightcoral!15, very thick, text=black
                            [
                            \cite{tam2023evaluating}{, }\cite{laban2023summedits}{, }\cite{tang2024tofueval}
                                    , leaf, text width=8em, color=lightcoral!100, fill=lightcoral!15, very thick, text=black
                            ]
                        ]
                        [
                        Position bias, color=lightcoral!100, fill=lightcoral!15, very thick, text=black
                            [
                            \cite{ravaut2023position}{, }\cite{chhabra2024revisiting}
                                    , leaf, text width=8em, color=lightcoral!100, fill=lightcoral!15, very thick, text=black
                            ]
                        ]
                        [
                        Fariness, color=lightcoral!100, fill=lightcoral!15, very thick, text=black
                            [
                            \cite{zhang2023fair}
                                    , leaf, text width=8em, color=lightcoral!100, fill=lightcoral!15, very thick, text=black
                            ]
                        ]
                    ]
                    [
                        Application, color=lightcoral!100, fill=lightcoral!15, very thick, text=black
                        [
                        Medical, color=lightcoral!100, fill=lightcoral!15, very thick, text=black
                            [
                            \cite{tang2023evaluating}{, }\cite{shaib2023summarizing}{, }\cite{van2023clinical}
                                    , leaf, text width=8em, color=lightcoral!100, fill=lightcoral!15, very thick, text=black
                            ]
                        ]
                        [
                        Code, color=lightcoral!100, fill=lightcoral!15, very thick, text=black
                            [
                            \cite{luo2023chatgpt}{, }\cite{sun2023automatic}{, }\cite{haldar2024analyzing}{, }\cite{haldar2024analyzing}
                                    , leaf, text width=8em, color=lightcoral!100, fill=lightcoral!15, very thick, text=black
                            ]
                        ]
                    ]
                ]
                [
                    Modeling (\S \ref{sec:method}), color=lightgreen!100, fill=lightgreen!15, very thick, text=black
                    [
                        Prompting, color=lightgreen!100, fill=lightgreen!15, very thick, text=black
                        [
                            CoT, color=lightgreen!100, fill=lightgreen!15, very thick, text=black
                            [
                                SumCoT~\cite{wang2023element}{, }CoD~\cite{adams2023sparse}{, }\\ChartThinker~\cite{liu2024chartthinker}, leaf, text width=8em, color=lightgreen!100, fill=lightgreen!15, very thick, text=black
                            ] 
                        ]
                        [
                            PromptSum~\cite{ravaut2023promptsum}{, }\cite{bhaskar2022zero}{, }\cite{chang2023booookscore}{, }\cite{sun2024prompt}, leaf, text width=13em, color=lightgreen!100, fill=lightgreen!15, very thick, text=black 
                        ]
                    ]
                    [
                        Multi-agent, color=lightgreen!100, fill=lightgreen!15, very thick, text=black
                        [
                            SummIt~\cite{zhang2023summit}{, }ImpressionGPT~\cite{ma2023impressiongpt}{, }\\ISQA~\cite{li2024isqa}{, }SliSum~\cite{li2024improving}, leaf, text width=13em, color=lightgreen!100, fill=lightgreen!15, very thick, text=black
                        ]
                    ]
                    [
                        Alignment, color=lightgreen!100, fill=lightgreen!15, very thick, text=black
                        [
                            \cite{liu2022improving}{, }SALT~\cite{yao2023improving}{, }InstructPTS~\cite{fetahu2023instructpts}, leaf, text width=13em, color=lightgreen!100, fill=lightgreen!15, very thick, text=black
                        ]
                    ]
                    [
                        Distillation, color=lightgreen!100, fill=lightgreen!15, very thick, text=black
                        [
                            \cite{liu2023learning}{, }\cite{jung2023impossible}{, }InheritSumm~\cite{xu-etal-2023-inheritsumm}{, }TriSum~\cite{jiang2024trisum}, leaf, text width=13em, color=lightgreen!100, fill=lightgreen!15, very thick, text=black
                        ]
                    ]
                    [
                        Training, color=lightgreen!100, fill=lightgreen!15, very thick, text=black
                        [
                            \cite{xia2024hallucination}{, }\cite{whitehouse2023parameter}{, }\cite{pu2024rst} , leaf, text width=13em, color=lightgreen!100, fill=lightgreen!15, very thick, text=black
                        ]
                    ]
                    [
                        Tool, color=lightgreen!100, fill=lightgreen!15, very thick, text=black
                        [
                            \cite{laskar2023can}{, }\cite{zhou2023multi}{, }\cite{mishra2023llm}{, }\cite{nakshatri2023using}{, }\cite{nawrath2024role}, leaf, text width=13em, color=lightgreen!100, fill=lightgreen!15, very thick, text=black
                        ]
                    ]
                ]
                [
                    Evaluation (\S \ref{sec:evaluation}), color=cyan!100, fill=cyan!15, very thick, text=black
                    [
                        General, color=cyan!100, fill=cyan!15, very thick, text=black
                        [
                                \cite{fu2023gptscore}{, }\cite{gao2023human}{, }\cite{jain2023multi}{, }\cite{liu2023gpteval}{, }\cite{wu2023large}{, }\cite{wu2023less}{, }\cite{shen2023large}
                                    , leaf, text width=13em, color=cyan!100, fill=cyan!15, very thick, text=black
                        ]
                    ]
                    [
                        Consistency, color=cyan!100, fill=cyan!15, very thick, text=black
                        [
                            \cite{luo2023chatgpt}{, }\cite{jia2023zero}{, }\cite{chen2023evaluating}{, }\cite{xu2024identifying}
                                    , leaf, text width=13em, color=cyan!100, fill=cyan!15, very thick, text=black
                        ]
                    ] 
                    [
                        Meta-eval, color=cyan!100, fill=cyan!15, very thick, text=black
                        [
                            \cite{chern2024can}
                                    , leaf, text width=13em, color=cyan!100, fill=cyan!15, very thick, text=black
                        ]
                    ]
                ]
            ]
        \end{forest}
}
    \caption{Taxonomy of Summarization Research with LLMs.}
    \label{fig:tax}
\end{figure*}
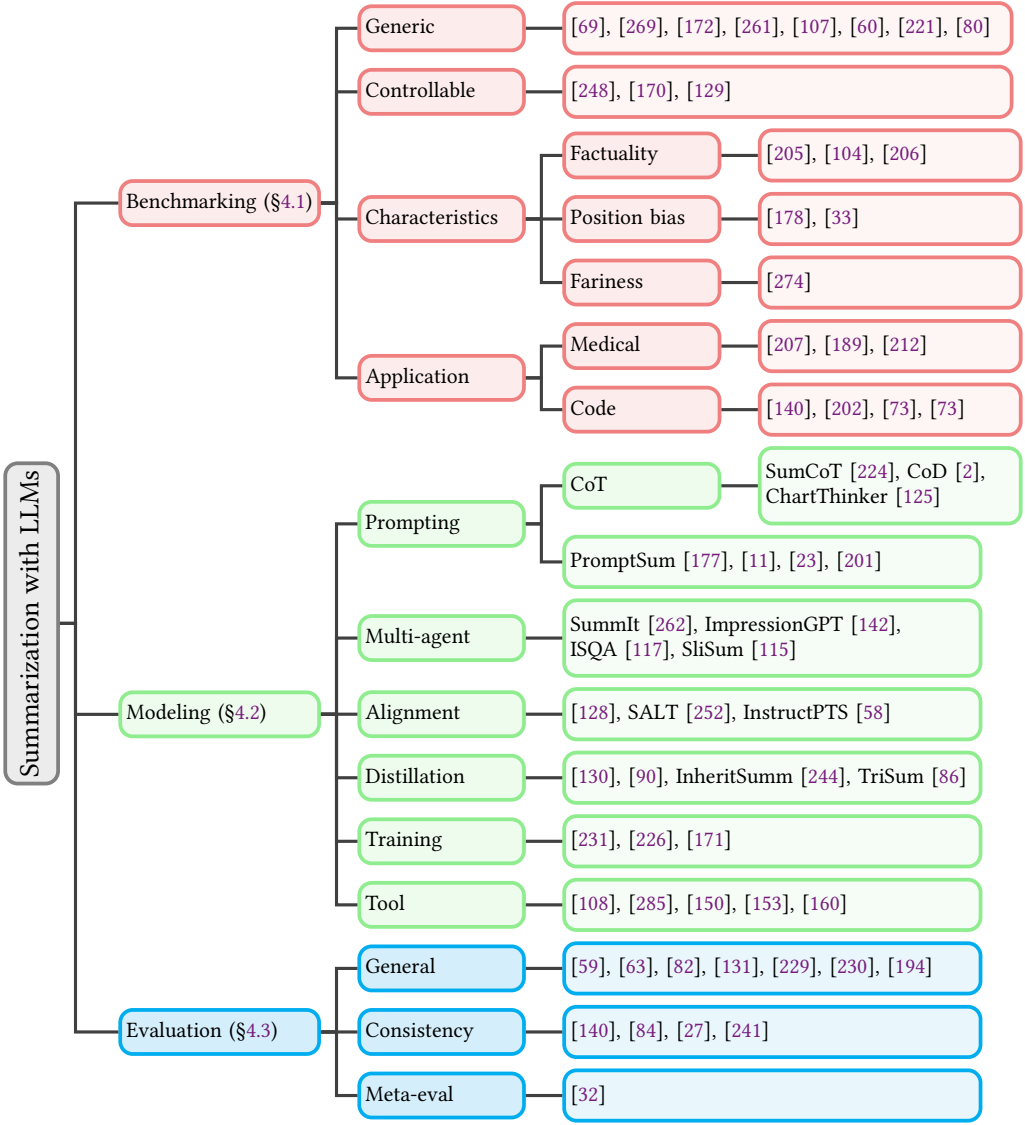

An overview of the current research landscape in LLM-based summarization is presented in Fig.~\ref{fig:tax}. Existing research works can be broadly categorized into three distinct types based on the objective of the study: LLM benchmarking studies, summarization modeling studies, and summary evaluation studies:

\begin{itemize}
    \item Benchmarking studies: These studies focus on evaluating the performance of LLMs across various summarization tasks. They aim to establish baselines and performance metrics that help in comparing different models and approaches. Benchmarking studies are crucial for understanding the strengths and limitations of LLMs in summarization and for guiding future research directions. They often involve extensive experimentation and analysis across diverse datasets and domains, providing a comprehensive performance landscape of existing models. Representative benchmarking studies are presented in Section~\ref{sec:benchmark} and summarized in Table~\ref{tab:llm_bench}.

    \item 

    Modeling Studies: This category encompasses research dedicated to developing novel summarization algorithms and architectures utilizing LLMs. These studies explore various techniques to enhance the summarization process, such as prompt engineering, multi-agent systems, model alignment, and knowledge distillation. Modeling studies aim to push the boundaries of what LLMs can achieve in summarization, focusing on improving the quality, coherence, and informativeness of generated summaries. They also investigate how different architectural modifications and training strategies can optimize the performance of LLMs for specific summarization tasks. Representative modeling studies are presented in Section~\ref{sec:method} and summarized in Table~\ref{tab:llm_method}.

    \item Evaluation Studies: These studies are concerned with devising new metrics and methods for assessing the quality and effectiveness of generated summaries. Unlike traditional metrics like ROUGE, evaluation studies in the era of LLMs focus on developing more sophisticated and human-aligned metrics. Representative evaluation studies are presented in Section~\ref{sec:evaluation} and summarized in Table~\ref{tab:llm_evaluation}.
\end{itemize}

\subsection{Benchmarking Studies}
\label{sec:benchmark}

With the advent of powerful new LLMs as backbone models, many research efforts are now aimed at benchmarking their performance and understanding their capabilities in the context of summarization. Existing works have explored benchmarking LLMs for generic and controllable summarization, examining characteristics such as faithfulness and position bias, and investigating their application in specialized domains such as medical and code summarization.

\subsubsection{Generic Summarization}

The emergence of GPT-3 marks a pivotal moment in the LLM era, demonstrating significant generative capabilities that spur a paradigm shift in NLP research. One of the pioneering benchmarking studies~\cite{goyal2022news} evaluates GPT-3 on text summarization, specifically within the classic benchmark domain of news summarization. This study compares GPT-3 against previously fine-tuned models and finds, through human evaluation, that annotators overwhelmingly preferred GPT-3's summaries, even when provided with only a task description. However, this preference is not reflected in automatic metrics, where GPT-3's summaries receive lower scores in both reference-based and reference-free evaluations.

Further large-scale human evaluation by \citet{zhang2023benchmarking} echoes these findings, highlighting limitations due to low-quality reference summaries. These limitations lead to underestimations of human performance and reduced few-shot and fine-tuning effectiveness. Their study underscores that LLM-generated summaries are judged to be comparable to human-written ones and emphasizes the significance of instruction tuning in enhancing the zero-shot summarization capabilities of LLMs. Similarly, human evaluation~\cite{pu2023summarization} reaches similar conclusions, arguing that LLM-generated summaries are noted for better factual consistency and fewer instances of extrinsic hallucinations.

For extractive summarization, \citet{zhang2023extractive} conduct a thorough evaluation of ChatGPT’s extractive performance against traditional fine-tuning methods. Their findings reveal that while ChatGPT underperforms in terms of ROUGE scores compared to existing supervised systems, it achieves higher performance based on LLM-specific evaluation metrics. The study also explores the benefits of in-context learning (ICL) and proposes an extract-then-generate pipeline to improve the faithfulness of LLM-generated summaries. Here ICL is a method of prompt engineering that allows LLMs to learn new tasks without fine-tuning by using examples as part of a prompt~\cite{dong2022survey}.

For meeting summarization, \citet{laskar2023building} further conduct an evaluation of LLMs on meeting summarization, discussing the cost-effectiveness and computational demands. This study provides a practical perspective on the trade-offs involved in using LLMs in real-world applications, highlighting the balance between performance and resource requirements. Additionally, \citet{fu2024tiny} investigate the viability of smaller, more compact LLMs as alternatives to their larger counterparts. They find that most smaller LLMs, even after fine-tuning, did not surpass larger zero-shot LLMs in summarization tasks.

In addition, the evaluation of LLMs in the multilingual setting~\cite{wang2023cross} reveal that ChatGPT and GPT-4 tend to produce detailed, lengthy summaries, and can achieve a better balance between informativeness and conciseness through interactive prompting. \citet{huang2023embrace} explore applying LLMs in multi-document summarization and point out that the task remains a complex challenge for LLMs mainly due to their limited coverage.

\subsubsection{Controllable/query-focused Summarization}

\citet{yang2023exploring} conduct an evaluation of ChatGPT's performance on query-based summaries, revealing that ChatGPT's performance is comparable to traditional fine-tuning methods in terms of ROUGE scores. \citet{liu2023benchmarking} evaluate LLMs on instruction-controllable text summarization, where the model input consists of both a source article and a natural language requirement for the desired summary characteristics. The study finds that instruction-controllable text summarization remains a challenging task for LLMs due to persistent factual errors in the generated summaries, a lack of strong alignment between LLM-based evaluation methods and human annotators when assessing the quality of candidate summaries, and performance discrepancies among different LLMs.

Additionally, \citet{pu2023chatgpt} conduct a systematic inspection of ChatGPT’s performance in adapting its output to different target audiences (experts vs. laymen) and writing styles (formal vs. informal). Their findings indicate that the stylistic variations produced by humans are considerably larger than those demonstrated by ChatGPT. Moreover, ChatGPT sometimes incorporates factual errors or hallucinations when adapting the text to suit a specific style.

\subsubsection{LLM Summarization Characteristics} Recent benchmarking studies have also explored the characteristics of LLMs when generating summaries, including generation faithfulness, fairness, and position bias. These studies aim to understand how accurately LLMs produce summaries that reflect the source content and how the position of information within the source text affects the generated summaries.

\textbf{Faithfulness and Factuality}: 
\citet{tam2023evaluating} propose a factual inconsistency benchmark to measure LLM factuality preferences in summarization. The benchmark includes human-verified factual summaries and annotated inconsistent summaries, providing a means to assess a model's factual consistency. The study finds that while existing LLMs generally prefer factually consistent summaries, they often fail to detect inconsistencies when content is copied verbatim.

\citet{tang2024tofueval} introduce an evaluation benchmark for topic-focused dialogue summarization, which includes human annotations of factual consistency along with detailed explanations. Their analysis reveals that LLMs often generate significant amounts of factual errors in the dialogue domain and perform poorly as binary factual evaluators. \citet{laban2023summedits} reach similar conclusions, finding that most LLMs struggle to detect factual inconsistency, with performance close to random chance. Additionally, they highlight that some existing competitive results are caused by issues with existing evaluation benchmarks. Overall, there are still gaps in LLMs’ ability to reason about facts and detect inconsistencies.

\textbf{Fairness}: \citet{zhang2023fair} define summary fairness as not underrepresenting perspectives of any groups of people and find that both the model-generated and the human-written reference summaries suffer from low fairness.

\textbf{Position bias}: Lead bias is a common phenomenon in news summarization, where the early parts of an article often contain the most salient information. \citet{ravaut2023position} conduct a comprehensive study on context utilization and position bias in summarization with LLMs, revealing that models tend to favor the initial and final segments of the input but largely ignore the middle, resulting in a U-shaped performance pattern. Similarly, \citet{chhabra2024revisiting} examine zero-shot abstractive summarization by measuring position bias, which refers to the tendency of a model to unfairly prioritize information from certain parts of the input text over others. Their study shows that LLMs exhibit significant lead bias in extreme summarization tasks.

\subsubsection{Applications} Recent research also explores applying LLMs in interdisciplinary domains like medical document processing and source code.

\textbf{Medical Document Summarization}: \citet{van2023clinical} use LLMs to summarize clinical documents, including radiology reports, patient questions, progress notes, and doctor-patient dialogue. Their clinical reader study finds that summaries from the best-adapted LLMs are deemed either equivalent or superior compared to those from medical experts. However, these LLM-generated summaries face safety challenges such as hallucinations and factual errors, which could potentially lead to medical harm.

In contrast, \citet{tang2023evaluating} apply LLMs in zero-shot medical evidence summarization across different clinical domains. They demonstrate that LLMs are susceptible to generating factually inconsistent summaries and making overly convincing or uncertain statements, which could lead to misinformation. Additionally, the models struggle to identify salient information and are more error-prone when summarizing longer texts.

\citet{shaib2023summarizing} enlist domain experts to evaluate summaries of biomedical articles generated by GPT-3 in both single- and multi-document settings. The study finds that while GPT-3 could faithfully summarize and simplify single biomedical articles, it struggled to provide accurate aggregations of findings across multiple documents.

\textbf{Code Summarization}: \citet{ahmed2022few} first examine the application of few-shot training with GPT Codex and demonstrate its ability to outperform state-of-the-art models for code summarization through the utilization of project-specific training. Subsequently, \citet{sun2023automatic} assess ChatGPT's performance on the extensively employed Python dataset, CSN-Python, revealing that its code summarization performance falls short in comparison to contemporary models in terms of BLEU and ROUGE-L metrics. 

Moreover, \citet{luo2023chatgpt} delve into the potential of LLMs in comprehending binary code. They discover that LLMs exhibit superior performance when applied to decompiled code in comparison to IR code, assembly code, and raw bytes. Notably, they identify that function names play a significant role in shaping the semantics of decompiled code. Furthermore, their research reveals that zero-shot prompts surpass few-shot and chain-of-thought prompts both in terms of performance and cost efficiency for large-scale tasks. Similarly, \citet{haldar2024analyzing} observe that the performance of models on individual instances often hinges on the subword token overlap between the code and corresponding natural language descriptions, particularly prevalent in function names.

\begin{table*}[!htbp]
  \caption{Overview of Benchmarking Studies on Summarization with LLMs.}
  \label{tab:llm_bench}
  \scalebox{0.90}{
\begin{tabular}{c|c|c|c|c}
\toprule
Source & Year &Task/Domain &LLM &Dataset \\

\midrule

\cite{goyal2022news}&2022&News&GPT3&CNN/DM, XSum, Newsroom\\\hline

\cite{zhang2024benchmarking}&2023&News&GPT3&CNN/DM, XSum\\\hline

\cite{pu2023summarization}&2023&News&GPT4&CNN/DM, Multi-News, Mediasum\\\hline

\cite{zhang2023extractive}&2023&Extractive&ChatGPT&CNN/DM, XSum, Reddit, PubMed\\\hline

\cite{laskar2023building}&2023&Meeting&multiple& AMI, ICSI, QMSUM\\\hline

\cite{fu2024tiny}&2024&Meeting&Multiple&In-Domain, QMSUM\\\hline

\cite{wang2023cross}&2023&Multilingual&Multiple&CLS\\\hline

\cite{huang2023embrace}&2023&Multi-doc&Multiple&DIVERSESUMM\\\hline

\cite{yang2023exploring}&2023&QFS&ChatGPT&CovidET, NEWTS, QMSum, SQuALITY\\\hline

\cite{pu2023chatgpt}&2023&QFS&ChatGPT&ELIFE\\\hline

\cite{liu2023benchmarking}&2023&Controllable&Multiple&INSTRUSUM\\\hline

\cite{tam2023evaluating}&2023&Factuality&Multiple&CNN/DM, XSum\\\hline

\cite{laban2023summedits}&2023&Factuality&Multiple&AggreFact, DialSummEva\\\hline

\cite{tang2024tofueval}&2023&Factuality, dialogue&Multiple&MediaSum, MeetingBank\\\hline

\cite{zhang2023fair}&2023&Fairness&Multiple&Claritin, US Election, Yelp, etc.\\\hline

\cite{ravaut2023position}&2023&Position bias&Multiple&MiddleSum\\\hline

\cite{chhabra2024revisiting}&2024&Position bias&Multiple&CNN/DM, XSum,
News, Reddit\\\hline

\cite{tang2023evaluating}&2023&Medical&ChatGPT&Alzheimer, Kidney, Skin, etc.\\\hline

\cite{shaib2023summarizing}&2023&Medical&GPT3&RCT\\\hline

\cite{van2023clinical}&2023&Medical&Multiple&ProbSum, MeQSum, ACI-Bench, etc.\\\hline

\cite{luo2023chatgpt}&2023&Code&Multiple&BinSum\\\hline

\cite{sun2023automatic}&2023&Code&ChatGPT&CSN-Python\\\hline

\cite{haldar2024analyzing}&2024&Code&CodeT5, Llama, PaLM 2&
CodeXGLUE\\\hline

\end{tabular}}
\end{table*}

\subsection{Modeling Studies}
\label{sec:method}

As reviewed in Section~\ref{sec:prior}, significant progress in text summarization has been achieved over the past decade, largely due to the paradigm shift towards deep learning methods and fine-tuning pre-trained language models. Most previous summarization research has focused on developing new models, algorithms, systems, or architectures and validating their effectiveness on standard datasets, referred to as modeling studies. Recent advancements in summarization research also introduce new modeling approaches with LLMs, primarily in zero-shot and few-shot settings. These studies encompass a variety of methods, including prompting-based, multi-agent-based, and alignment-based techniques to enhance the quality of generated summaries. Additionally, distillation-based methods and innovative training strategies have been proposed to improve model efficiency. Furthermore, LLMs are increasingly utilized as tools to enhance system performance in text summarization tasks.

\subsubsection{Prompting-based Method}

PromptSum~\cite{ravaut2023promptsum} proposes a method that combines prompt tuning with a multi-task objective and discrete entity prompts for abstractive summarization. This approach employs a soft prompt to generate an entity chain, which is then used alongside another soft prompt for summary generation, thereby enhancing model efficiency and controllability. Subsequently, \citet{bhaskar2022zero} investigate prompting pipeline methods for long-form opinion summarization, including hierarchical summarization with chunking and pre-extraction using an extractive summarization model through a topic clustering-chunking-generation pipeline. Additionally, \citet{chang2023booookscore} introduce a prompting method to hierarchically merge chunk-level summaries and incrementally update a running summary, focusing on the application of LLMs for book-length summaries. \citet{sun2024prompt} study prompt chaining that performs drafting, critiquing, and refining phases through a series of three discrete prompts versus a stepwise prompt that integrates these phases within a single prompt.

\textbf{Chain of Thought}: Chain-of-thought (CoT) methods have demonstrated significant potential in eliciting the reasoning capabilities of LLMs, thereby enhancing the quality of generated summaries~\cite{wei2022chain}. Building on this, SumCoT~\cite{wang2023element} expands CoT into an element-aware summarization approach that prompts the LLM to first enumerate important facts and subsequently integrate these facts into a coherent summary, ensuring that the final output is both comprehensive and well-organized. Furthermore, chain-of-density (CoD)~\cite{adams2023sparse} iteratively incorporates missing salient entities into initially entity-sparse summaries while maintaining a fixed length, resulting in more informative summaries with reduced lead bias. Additionally, ChartThinker~\cite{liu2024chartthinker} proposes to synthesize deep analysis based on CoT and strategies of context retrieval, aiming to improve the logical coherence and accuracy of the generated summaries.

\subsubsection{Multi-agent based Method}

Recent research has explored the construction of multi-agent systems utilizing LLMs to enhance summarization quality. SummIt~\cite{zhang2023summit} introduces an iterative text summarization framework based on ChatGPT, incorporating multiple LLM agents such as summarizers and evaluators. This system refines the generated summaries through a process of self-evaluation and feedback, and also investigates the benefits of integrating knowledge and topic extractors to improve the faithfulness and controllability of the summaries. Similarly, ImpressionGPT~\cite{ma2023impressiongpt} designs an iterative optimization algorithm to perform automatic evaluations of the generated impressions and formulates corresponding instruction prompts to further refine the model's output. This approach leverages the in-context learning capabilities of LLMs by creating dynamic contexts with domain-specific, individualized data.

Moreover, ISQA~\cite{li2024isqa} proposes iterative factuality refining for informative scientific question-answering (ISQA) feedback. It operates in a detailed manner, instructing the summarization agent to reinforce validated statements from positive feedback and correct inaccuracies highlighted in negative feedback. SliSum~\cite{li2024improving} employs sliding windows and self-consistency with LLMs. It divides the source article into overlapping windows, allowing LLMs to generate local summaries for each window's content and then aggregates them using clustering and a majority voting algorithm.

\subsubsection{Alignment}
Recent advancements in summarization techniques have focused on leveraging feedback mechanisms and instruction-based fine-tuning to improve the quality and user alignment of generated summaries~\cite{ouyang2022training}. \citet{liu2022improving} first investigate the use of informational feedback from human annotations in natural language to enhance generation quality and factual consistency. Subsequently, SALT~\cite{yao2023improving} introduces a method that integrates both human-edited and model-generated data within the training loop by simulating human edits. InstructPTS~\cite{fetahu2023instructpts} proposes using instruction fine-tuning of LLMs to achieve controllable product title summarization across different dimensions, utilizing automatically generated instructions derived from a parallel dataset.

\subsubsection{Distillation-based Method}
Despite the strong summarization performance of LLMs, their resource demands have limited their widespread use. Additionally, privacy concerns have arisen when using LLM-as-a-service APIs, particularly for sensitive data. This underscores the need for more compact local models that can still effectively capture summarization abilities.

\citet{liu2023learning} proposed training smaller models like BART or BRIO using contrastive learning, where LLMs such as ChatGPT serve as evaluators to signal which generated summary candidate is superior. TriSum~\cite{jiang2024trisum} first extracts a set of aspect-triple rationales and summaries with LLMs, and then trains a smaller local model with a curriculum learning strategy on these tasks. Triple here is a
structure formatting a piece of free text into a subject, a relation, and an object.

Additionally, InheritSumm~\cite{xu-etal-2023-inheritsumm} achieves a compact summarization model with versatile capabilities across different settings by training using knowledge distillation from the GPT-3 model, mimicking the GPT-generated summaries on general documents. Similarly, \citet{jung2023impossible} present Impossible Distillation, leveraging the paraphrastic proximity inherent in pre-trained language models like GPT-2, identifying and distilling high-quality paraphrases from these subspaces.

\subsubsection{Training Strategy}

Recent research also explores training strategies to fine-tune LLMs. For example, \citet{xia2024hallucination} propose a method to alleviate hallucinations in LLM outputs using hallucination diversity-aware active learning. This approach selects diverse hallucinations for annotations in the active learning process for LLM fine-tuning.

\citet{whitehouse2023parameter} investigate the potential of parameter-efficient fine-tuning (PEFT) with low-rank adaptation (LoRA)~\cite{hu2021lora} in the domain of multilingual summarization. Furthermore, \citet{pu2024rst} propose to integrate rhetorical structure theory (RST) with LoRA to enhance long document summarization performance.

\subsubsection{Others}

Recent research efforts have also explored innovative approaches to leverage LLMs as a tool for summarization tasks, such as data cleaning for QFS summarization in~\cite{laskar2023can}, data annotation for pre-training for dialogue summarization in~\cite{zhou2023multi}, and pseudo-label generation for dialogs to fine-tune smaller model in~\cite{mishra2023llm}.

\begin{table*}[!htbp]
  \caption{Overview of LLM-based Summarization Methods.}
  \label{tab:llm_method}
  \scalebox{0.90}{
\begin{tabular}{c|c|c|c|c}
\toprule

Source & Year &Task/Domain &LLM &Dataset \\

\midrule

\cite{ravaut2023promptsum}&2023&Book&PEGASUS&CNN/DM, XSum, BillSum, SAMSum\\\hline

\cite{bhaskar2022zero}&2022&Opinion&GPT3.5&SPACE, FewSum\\\hline

\cite{chang2023booookscore}&2023&Book&GPT4, Claude 2, etc,&-\\\hline

\cite{sun2024prompt}&2024&News&GPT 3.5/4&InstruSum\\\hline

\cite{wang2023element}&2023&News&GPT3&CNN/DM, XSum\\\hline

\cite{adams2023sparse}&2023&News&GPT4&CNN/DM\\\hline

\cite{liu2024chartthinker}&2024&Chart&LLaMa2&Chart-Sum-QA\\\hline

\cite{zhang2023summit}&2023&News&ChatGPT&CNN/DM, XSum, NEWTS\\\hline

\cite{li2024isqa}&2024&Scientific&Llama2, Vicuna, Flan-T5&SciMRC, QASPER\\\hline

\cite{ma2023impressiongpt}&2023&Radiology report&ChatGPT&MIMIC-CXR, OpenI\\\hline

\cite{li2024improving}&2024&multiple&Llama2, Claude-2, GPT-3.5&CNN/DM, XSum, arXiv, PubMed\\\hline

\cite{liu2022improving}&2022&News&GPT3.5, T5&DEFACTO\\\hline

\cite{yao2023improving}&2023&Conversation&GPT2&CC\\\hline

\cite{fetahu2023instructpts}&2023&Title&Flan T5&-\\\hline

\cite{liu2023learning}&2023&News&ChatGPT,GPT4&CNN/DM, 
XSum\\\hline

\cite{xu-etal-2023-inheritsumm}&2023&Multiple&GPT-3.5, ZCode++&MultiNew, SAMSum, etc.\\\hline

\cite{jung2023impossible}&2023&Multiple&ChatGPT, GPT3.5, T5&Gigaword \\\hline

\cite{jiang2024trisum}&2023&News, clinical&GPT3.5&CNN/DailyMail, XSum, ClinicalTrial\\\hline

\cite{xia2024hallucination}&2024&news&Flan-T5, BART&CNN/DM, MultiNews, Gigaword\\\hline

\cite{pu2024rst}&2024&Long doc&Vicuna, GPT4&Multi-LexSum, eLife, BookSum \\\hline

\cite{laskar2023can}&2023&QFS&GPT4&Debatepedia\\\hline

\cite{zhou2023multi}&2023&Dialogue&ChatGPT&SAMSum, DIALOGSUM\\\hline

\cite{nakshatri2023using}&2023&News&GPT3.5&KEYEVENTS\\\hline

\cite{mishra2023llm}&2023&Dialogue&GPT3.5&Tweetsumm\\\hline

\hline
\end{tabular}}
\end{table*}

\subsection{Evaluation Studies}
\label{sec:evaluation}

Evaluating the quality of generated summaries accurately has indeed been a major challenge in summarization research. As discussed in Section~\ref{metric}, researchers have explored reference-based and reference-free automatic metrics, but the correlation with human judgment is still limited and cannot reliably evaluate summary quality~\cite{fabbri2021summeval}. With the recent emergence of LLMs, researchers are also investigating their use in summary quality evaluation.

\subsubsection{General Summary Evaluation}

GPTScore~\cite{fu2023gptscore} represents a pioneering effort in utilizing LLMs for summary evaluation metrics. It leverages the emergent abilities of LLMs by incorporating task specifications and detailed aspect definitions, which serve as instructions for scoring generated texts. Building on this, \citet{gao2023human} investigate ChatGPT's capacity for human-like summarization evaluation. Their findings indicate that ChatGPT can perform annotations relatively smoothly using various methods, including Likert scale scoring, pairwise comparison, the Pyramid method, and binary factuality evaluation.

Similarly, \citet{jain2023multi} propose using LLMs as multi-dimensional evaluators for text summarization through in-context learning, assessing aspects such as fluency, coherence, and factuality. G-Eval~\cite{liu2023gpteval}, another notable LLM-based evaluation metric for summaries, uses a chain-of-thought and form-filling paradigm. G-Eval demonstrates a high Spearman correlation with human evaluations on summarization tasks, significantly outperforming previous methods. \citet{wu2023large} introduce role-players prompting mechanism to evaluate summaries from both objective and subjective perspectives. Another approach proposed by \citet{wu2023less} aims to reduce evaluation costs with an Extract-then-Evaluate method.

Despite these advancements, \citet{shen2023large} investigate the stability and reliability of LLMs as automatic evaluators for abstractive summarization. They find that although models like ChatGPT and GPT-4 outperform traditional automatic metrics, they are not yet reliable substitutes for human evaluators. LLM evaluators exhibit inconsistent ratings across different candidate systems and dimensions, struggle with close performance comparisons, and show a lower correlation with human judgments as the quality of summaries improves.

\subsubsection{Factual Consistency Evaluation}

\citet{luo2023chatgpt} first explore ChatGPT's ability to evaluate factual inconsistency under a zero-shot setting. They find that ChatGPT achieves satisfying performance but prefers more lexically similar candidates and instances of false reasoning. \citet{chen2023evaluating} further extend the setting with different LLMs and different prompting techniques.

Subsequently, \citet{jia2023zero} propose a zero-shot faithfulness evaluation with a foundation language model without prompting or instruction tuning. \citet{xu2024identifying} propose three zero-shot paradigms for factual consistency evaluation, including direct inference on the entire summary or each summary window, and entity verification through question generation and answering.

\subsubsection{Others}

\citet{chang2023booookscore} propose an automatic metric BOOOOKSCORE to assess summary coherence in book-length summary generation. \citet{huang2023embrace} explore the usage of automatic LLM-based metrics under the multi-document summarization setting.

In addition, \citet{chern2024can} conduct a meta-evaluation that can effectively, reliably, and efficiently evaluate the performance of LLMs as evaluators across diverse tasks and scenarios via agent debate.

\begin{table*}[!htbp]
  \caption{Overview of LLM-based Summary Evaluation.}
  \label{tab:llm_evaluation}
  \scalebox{0.90}{
\begin{tabular}{c|c|c|c}
\toprule
Source & Year &Aspect &LLM \\

\midrule

\cite{fu2023gptscore}&2023&Coverage, Factuality, Coherence, Informativeness, etc.&GPT3, OPT, FLAN-T5, GPT2\\\hline

\cite{gao2023human}&2023&Coverage, Consistency, Coherence, Fluency, etc.&ChatGPT\\\hline

\cite{jain2023multi}&2023&Coherence, Consistency, Fluency, Relevance&GPT3\\\hline

\cite{liu2023gpteval}&2023&Coverage, Factuality, Coherence, Informativeness, etc.&GPT4\\\hline

\cite{wu2023large}&2023&General&GPT3\\\hline

\cite{wu2023less}&2023&Consistency, Relevance, Faithfulness&GPT4\\\hline

\cite{shen2023large}&2023&Coherence, Consistency, Fluency, Relevance&GPT4, ChatGPT\\\hline

\cite{luo2023chatgpt}&2023&Consistency&ChatGPT\\\hline

\cite{jia2023zero}&2023&Consistency& LLaMa\\\hline

\cite{chen2023evaluating}&2023&Consistency&Multiple\\\hline

\cite{xu2024identifying}&2024&Consistency&Llama-2, Vicuna, ChatGPT\\\hline

\cite{chern2024can}&2024&Meta-evaluation&GPT3.5, Claude, Gemini, etc.\\\hline

\hline
\end{tabular}}
\end{table*}

\section{Open Problems and Future Directions}
\label{sec:future}

This section discusses the trends in recent LLM-based summarization,  the general challenges of text summarization that have yet to
be solved and pinpoints potential future research directions to attract practitioners’ attention
and improve our understanding and techniques in text summarization in the new era of LLMs.

\subsection{Quantitative Results}

The CNN/DM dataset~\cite{hermann2015teaching} has been used as the predominant benchmark for assessing summary quality and comparing the efficacy of summarization methodologies in recent years. Here, we also review the quantitative results of representative summarization approaches on it, as delineated in Table~\ref{tab:result}, to glean insights into the trajectory of research advancements. Despite the limitations discussed in Section~\ref{metric}, ROUGE F-scores remain the standard and most widely adopted way for the assessment of summary quality, so we report ROUGE scores in the table for comprehensive evaluation.

The quantitative results in Table~\ref{tab:result} reveal significant advancements in summarization systems over recent years, driven by paradigm shifts towards deep learning and PLM-based approaches, which have notably elevated ROUGE performance. Despite CNN/DM's establishment as a predominantly extractive dataset~\cite{zhong2019closer}, abstractive summarization systems have demonstrated superior flexibility and performance compared to extractive methods. However, it's worth noting that ROUGE scores may inadequately capture summary quality, particularly in the new era of LLMs and when reference summaries are of low quality, as discussed in~\cite{goyal2022news,zhang2024benchmarking}. Consequently, recent research has shifted focus towards human evaluations to more accurately assess summarization system performance in the context of evolving language models.

\begin{table}[!htbp]
  \caption{Quantitative Results on the CNN/DM dataset. Results are taken from the original paper if available.}
  \label{tab:result}
\begin{tabular}{c|c|c|c|c|c}
\toprule
Method&Model &Year&ROUGE-1 &ROUGE-2 &ROUGE-L\\

\midrule
Extractive(unsup.)&TextRank\cite{mihalcea2004textrank}&2004&33.2& 11.8 &29.6\\

Extractive&SUMMARUNNER~\cite{nallapati2017summarunner}&2017&39.60 &16.20 &35.30\\

Extractive&REFRESH~\cite{narayan2018ranking}&2018 &40.00 &18.20 &36.60\\

Extractive&BANDITSUM~\cite{dong2018banditsum}&2018 &41.50 &18.70 &37.60\\

Extractive&NEUSUM~\cite{zhou2018neural} &2018&41.59 &19.01 &37.98\\

Extractive (unsup.)&PACSUM~\cite{zheng2019sentence}&2019&40.7& 17.8& 36.9\\

Extractive&HIBERT~\cite{zhang2019hibert} &2019&42.37 &19.95 &38.83\\

Extractive&BERTSUMEXT~\cite{liu2019text} &2019&43.85 &20.34 &39.90\\
Extractive&DISCOBERT~\cite{xu2019discourse} &2019&43.77 &20.85 &40.67\\

Extractive&HSG~\cite{wang2020heterogeneous} &2020&42.95 &19.76 &39.23\\

Extractive&MATCHSUM~\cite{zhong2020extractive}&2020&44.41 &20.86& 40.55\\

Extractive&DiffuSum~\cite{zhang2023diffusum}&2023 &44.83& 22.56& 40.56\\
\hline

Abstractive&PTGEN~\cite{see2017get}&2017 &36.44 &15.66 &33.42\\

Hybrid&BOTTOMUP~\cite{gehrmann2018bottom}&2018 &41.22 &18.68 &38.34\\

Abstractive&DCA~\cite{celikyilmaz2018deep}  &2018&41.69 &19.47& 37.92\\

Abstractive&BERTSUMABS~\cite{liu2019text} &2019&41.72& 19.39& 38.76\\

Abstractive&BART~\cite{lewis2019bart}&2019&44.16 &21.28& 40.90\\
Abstractive&PEGASUS~\cite{zhang2020pegasus}&2020& 44.17& 21.47& 41.11\\
Hybrid&GSum~\cite{dou2020gsum}&2020&45.94& 22.32 &42.48\\
  
Abstractive&SimCLS~\cite{liu2021simcls}&2021&46.67 &22.15 &43.54\\

\hline

Abstractive&SummIt (ChatGPT)~\cite{zhang2023summit}&2023&37.29 &13.60  &26.87\\
Abstractive&Element (GPT3)~\cite{wang2023element}&2023&37.75 &15.20 &34.25\\

Abstractive&SliSum (Claude2)~\cite{li2024improving}&2024&47.75 &23.16 &44.26\\

Abstractive&HADAS ~\cite{xia2024hallucination}&2024&- &- &20.12\\

\hline
\end{tabular}
\end{table}

\subsection{Research Trends}
\label{sec:finding}

Based on the investigation and taxonomy of existing LLM-based summarization research discussed in Secton~\ref{sec:llm}, we summarize some research trends in the existing literature.

\textbf{LLM-based summarization is still nascent:} Although significant research efforts have been dedicated to LLM-based summarization, we are still in the early stages of fully leveraging LLMs for these tasks. Most existing research focuses on benchmarking studies that evaluate the performance of LLMs across various summarization tasks and analyze the behavior and characteristics of these 'black box' tools. This contrasts with the much more advanced modeling studies seen during the deep learning and PLM fine-tuning stages, which propose methods to better utilize these models for building summarization systems. The recent availability of open-source LLMs, such as LLaMa \cite{touvron2023llama}, has facilitated this kind of research. The next focus will be on developing more effective LLM-based summarization systems for diverse scenarios and enhancing the efficacy of LLMs in these tasks.

\textbf{Summarization is expanding horizons:}
While the majority of summarization research during the deep learning and PLM stages heavily focused on specific domains like news and dialogue, the attention is now shifting towards exploring broader domains and applications, including interdisciplinary topics. Researchers are investigating ways to create more holistic and versatile systems, opening up new avenues for their applications in fields such as finance \cite{oyewole2024automating}, law \cite{deroy2023ready, katz2023natural}, healthcare \cite{wang2023prompt, joshi2020dr}, and beyond \cite{chen2024survey}.

\textbf{The growing role of human involvement:}
The role of human involvement in summarization research is becoming increasingly crucial. Researchers now heavily rely on human evaluation to assess the quality of summarization outputs due to the limitations of automatic metrics~\cite{fabbri2021summeval,liu2023gpteval}. Human feedback is vital for refining models to better meet user needs. Moreover, hybrid approaches that combine human expertise with LLM capabilities are being explored to enhance the quality and reliability of summaries~\cite{stiennon2020learning,wu2021recursively}. Human-in-the-loop systems, where human feedback is continuously used to improve model performance, are gaining traction~\cite{chen2022human, shapira2022interactive}.

\textbf{Summarization is transitioning to real-world applications:}
With the paradigm shift towards LLM-based summarization, new research opportunities have emerged, positioning the field on the brink of significant growth and real-world application~\cite{chen2024survey}. The increasing demand for efficient information processing tools across various industries—such as news, legal, healthcare, and education—drives the development of robust summarization systems for practical use. Current models like GPT-4 have demonstrated impressive capabilities in generating coherent and contextually appropriate summaries, enabling the practical deployment of these systems~\cite{laskar2023building}. Real-world applications are beginning to emerge, and we can expect a proliferation of LLM-based summarization tools tailored for specific domains, thereby enhancing productivity and decision-making processes.

\subsection{Open Challenges}
Despite the significant progress made in recent years, there are still open challenges in text summarization research in the new era of LLMs.

\subsubsection{Hallucination:} Hallucination is one of the most significant issues hindering LLMs from real-world deployments. It refers to instances where the model generates information that is factually incorrect or not present in the source text, including fabricating facts, events, or details that were never mentioned in the original document. This issue is particularly problematic in critical fields such as medicine and law, where accuracy is paramount. Researchers have explored various approaches to measure and quantify the level of hallucination through benchmarks~\cite{lin2021truthfulqa}, and to mitigate hallucination in LLM generation with enhanced training protocols~\cite{zhang2023siren, ji2023survey}, retrieval-augmented generation (RAG)~\cite{lewis2020retrieval, chen2024benchmarking, yu2023improving, li2024unveiling}, self-reflection~\cite{shinn2023reflexion, zhang2023summit}, and post-processing techniques~\cite{chen2023purr, chen2021improving}. However, more efforts are still needed to achieve faithful and factual summary generation.

\subsubsection{Bias:}
Large language models can reflect and amplify biases present in their training data, leading to the generation of biased summaries~\cite{zhang2023fair, navigli2023biases, gallegos2023bias}. Addressing these ethical issues involves developing methods to detect bias in generation and mitigate biased content in the summary generation process. There is a growing need for models that produce fair and unbiased summaries, particularly in socially sensitive domains.

\subsubsection{Computational Efficiency:}
The computational efficiency of LLMs is a critical factor influencing their performance, scalability, and practical application. These models require significant computational resources due to their extensive parameter sizes and the complexity of training and inference. Efficiency improvements can be achieved through hardware acceleration~\cite{yang2020retransformer, yang2022building} and techniques like model pruning~\cite{sun2023simple, ma2023llm}, quantization~\cite{xiao2023smoothquant, li2023loftq}, parameter-efficient fine-tuning (PEFT)~\cite{hu2021lora, dettmers2024qlora}, and knowledge distillation~\cite{gu2023minillm, xu2024survey, jung2023impossible}. Despite these advancements, balancing computational efficiency with model accuracy and generalization remains a challenging and ongoing area of research.

\subsubsection{Personalization}:
Creating summaries personalized to individual user preferences poses another significant challenge~\cite{kirk2024benefits, wu2023tidybot}. This task involves understanding user interests, reading habits, and prior knowledge to generate summaries that are relevant and useful to the specific user. It requires NLP techniques that not only condense information but also align it with the user's unique profile. Personalized summarization can be particularly useful in fields such as education, news, and media, where it can adapt content to different learning styles and present information that aligns with the reader's interests~\cite{porsdam2023autogen, cai2023generating, cheng2023towards}. However, this personalization must be balanced with privacy considerations, ensuring that user data is handled securely and ethically.

\subsubsection{Interpretability and explainability}: Interpretability and explainability of LLM-based summarization are crucial for building trust and ensuring transparency in AI-generated content~\cite{lyu2024towards}. Interpretability refers to the ability to understand how a model makes its decisions, while explainability involves elucidating the reasons behind specific outputs. In the context of summarization, these aspects help users comprehend why certain information was highlighted or omitted~\cite{DBLP:journals/corr/abs-2112-05364, sarkhel2020interpretable}. Providing clear explanations for summarization outputs is particularly important in sensitive domains like healthcare and legal fields, where understanding the rationale behind the summary can impact decision-making processes~\cite{chen2024survey}. Despite advancements, achieving high levels of interpretability and explainability remains challenging due to the complexity and black-box nature of LLMs.

\subsection{Future Directions}
Here, we also highlight some future directions for summarization in the era of LLMs to broaden the scope of existing summarization systems, thereby expanding their application scenarios and impacts.

\subsubsection{Summarization beyond text}
Over the past few decades, significant progress has been made in the field of text summarization research. The advent of powerful LLMs such as GPT-4 and GPT-4o~\cite{achiam2023gpt}, capable of processing various data formats, has catalyzed advancements in multimodal summarization. This emerging area extends traditional text summarization to include other forms of data such as images~\cite{zhu2018msmo, qu2020context,zhu2020multimodal, mukherjee2022topic,xiao2023cfsum,jiang2023exploiting}, tables~\cite{liu2023neural}, source code~\cite{wan2018improving,wang2020reinforcement}, audio~\cite{kano2023speech}, and video~\cite{lin2023videoxum, krubinski2023mlask}. Multimodal summarization systems aim to provide more comprehensive and contextually rich summaries by integrating diverse data sources. These systems enable the generation of summaries that incorporate key visual elements from videos or important audio clips from podcasts, thereby enhancing user understanding and expanding real-world applications. This holistic approach has the potential to revolutionize information consumption, making it more accessible and engaging across various formats, and represents a significant step towards the development of AI assistants.

\subsubsection{New tasks in summarization}

Traditional summarization research has heavily focused on generic and extractive approaches. However, with the advent of powerful LLMs, the horizon of text summarization is brimming with the promise of tackling novel tasks that extend beyond conventional settings. Emerging research is steering toward more nuanced aspects tailored to specific user scenarios and real-world applications.

These include personalized and customized summarization that aligns with individual preferences, interests, and reading history~\cite{zhong2022unsupervised,cheng2023towards}, human-in-the-loop summarization enabling user engagement and verification for high-stakes domains like legal and medical fields~\cite{avinesh2018sherlock,shapira2021extending,shapira2017interactive}, real-time summarization that enhances the speed and efficiency of models for live events or breaking news~\cite{zhao2014quasi,sequiera2018overview,yang2022catchlive,cho2021streamhover}, sentiment-aware summarization reflecting the underlying sentiment of source material~\cite{xu2023sentiment,yang2018aspect}, and diversified summarization consolidating information across multiple articles~\cite{huang2023embrace}, and beyond. These new frontiers not only broaden the scope of summarization tasks but also hold immense potential in advancing the sophistication and utility of LLM-generated summaries across various domains and applications.

\subsubsection{Ethical and responsible studies}

Developing ethical AI practices is crucial for the future of summarization~\cite{gallegos2023bias, liang2021improving}. This includes creating transparent models that can explain their summarization processes, ensuring fairness in summaries that do not underrepresent perspectives of any groups of people, and actively working to mitigate biases~\cite{zhang2023fair, hsu2021decision}. There is also an urgent need for more theoretical groundings and ethical considerations for model generation from cognitive science, social science, psychology, and linguistics. Effectively and safely incorporating updated world knowledge and common sense into the summary generation cycle with control is also important to ensure the summaries are accurate, up-to-date, and factually correct~\cite{chen2021improving}. Ethical considerations will be paramount in building trust and ensuring the responsible use of summarization technologies.

\subsubsection{Domain Specific Summarization LLM}

Current LLMs demonstrate extraordinary generalization capability across different domains and subjects, thanks to large-scale pre-training with web data. However, this generalization capability may limit their practical use in specific domains when operating in the zero-shot setting due to the lack of domain-specific knowledge, rules, patterns, and terminologies. Consequently, the need for domain-specific LLM-based summarization systems is growing and necessary for practical deployment~\cite{kasneci2023chatgpt, thirunavukarasu2023large, cui2023chatlaw, xu2022systematic}. Effectively and efficiently adapting open-sourced LLMs to domain-specific applications is an important direction in building trustworthy expert summarization systems.
\section{Conclusion}

In conclusion, the landscape of text summarization research has witnessed profound advancements driven by deep neural networks, PLMs, and the recent emergence of LLMs. This survey captures the evolution of text summarization paradigms in detail, offering a comprehensive examination of both pre-LLM and LLM eras. Furthermore, the survey highlights research trends and delineates open challenges, proposing future research directions crucial for advancing summarization techniques. Through this comprehensive synthesis, the survey aims to serve as a valuable resource for researchers and enthusiasts, facilitating informed exploration and innovation in the ever-evolving domain of text summarization.


\bibliographystyle{ACM-Reference-Format}
\bibliography{sample-base}


\end{document}